\documentclass{article} 
\usepackage{iclr2017_conference,times}
\usepackage{hyperref}
\usepackage{url}
\usepackage{algorithm}
\usepackage{algorithmic}
\usepackage{stackrel}
\usepackage{amsmath}
\usepackage{amssymb}
\usepackage{subfigure}
\usepackage{graphicx}

\title{Learning Locomotion Skills Using DeepRL:\\ Does the Choice of Action Space Matter?}

\author{Xue Bin Peng \& Michiel van de Panne \\
Department of Computer Science \\
University of the British Columbia \\
Vancouver, Canada \\
\texttt{\{xbpeng,van\}@cs.ubc.ca} \\
}

\begin{document}

\maketitle

\begin{abstract}

The use of deep reinforcement learning allows for high-dimensional state descriptors, but little is known about how the choice of 
action representation impacts the learning difficulty and the resulting performance.
We compare the impact of four different action parameterizations (torques, muscle-activations, target joint angles, and target joint-angle velocities)
in terms of learning time, policy robustness, motion quality, and policy query rates.
Our results are evaluated on a gait-cycle imitation task for multiple planar articulated figures and multiple gaits.
We demonstrate that the local feedback provided by higher-level action parameterizations can significantly impact the learning, robustness, and quality of the resulting policies.
\end{abstract}

\section{Introduction}
The introduction of deep learning models to reinforcement learning (RL) has enabled policies to operate 
directly on high-dimensional, low-level state features. 
As a result, deep reinforcement learning (DeepRL) has demonstrated impressive capabilities,
such as developing control policies that can map from input image pixels to output joint torques ~\citep{DBLP:journals/corr/LillicrapHPHETS15}. 
However, the quality and robustness often falls short of what has been achieved with hand-crafted action abstractions, e.g., \citet{2011-TOG-quadruped,2013-TOG-MuscleBasedBipeds}. Relatedly,  the {\em choice of action parameterization} 
is a design decision whose impact is not yet well understood.

Joint torques can be thought of as the most basic and generic representation for driving 
the movement of articulated figures, given that muscles and other actuation models eventually result 
in joint torques.
However this ignores the intrinsic embodied nature of biological systems, particularly the synergy between control and biomechanics.
Passive-dynamics, such as elasticity and damping from muscles and tendons, 
play an integral role in shaping motions: they provide mechanisms for energy storage, and 
mechanical impedance which generates instantaneous feedback without requiring 
any explicit computation. Loeb coins the term {\em preflexes}~\citep{loeb1995control} to describe 
these effects, and their impact on motion control has been described 
as providing {\em intelligence by mechanics}~\citep{blickhan2007intelligence}.

In this paper we explore the impact of four different actuation models on learning to control
dynamic articulated figure locomotion: (1) torques (Tor); (2) activations for musculotendon units
(MTU); (3) target joint angles for proportional-derivative controllers (PD); and (4) target joint
velocities (Vel). Because Deep RL methods are capable of learning control policies for all these
models, it now becomes possible to directly assess how the choice of actuation model affects the learning
difficulty.  We also assess the learned policies with respect to robustness, motion
quality, and policy query rates.  We show that action spaces which incorporate local feedback can
significantly improve learning speed and performance, while still preserving the generality afforded
by torque-level control. Such parameterizations also allow for more complex body structures
and subjective improvements in motion quality.


Our specific contributions are: (1) We introduce a DeepRL framework for motion imitation tasks; 
(2) We evaluate the impact of four different actuation models on learned control policies according to four criteria; and 
(3) We propose an optimization approach that combines policy learning and 
actuator optimization, allowing neural networks to effective control complex muscle models.

\section{Background}
Our task will be structured as a standard reinforcement problem where an agent interacts with its environment according to a policy in order to maximize a reward signal. The policy $\pi(s, a) = p(a|s)$ represents the conditional probability density function of selecting action $a \in \mathcal{A}$ in state $s \in \mathcal{S}$. At each control step $t$, the agent observes a state $s_t$ and samples an action $a_t$ from $\pi$. The environment in turn responds with a scalar reward $r_t$, and a new state $s'_t = s_{t+1}$ sampled from its dynamics $p(s' | s, a)$. For a parameterized policy $\pi_\theta(s, a)$, the goal of the agent is learn the parameters $\theta$ which maximizes the expected cumulative reward
\[ J(\pi_\theta) = \mathop{\mathbb{E}}\left[\sum_{t=0}^T \gamma^t r_t \middle| \pi_\theta \right]  \]

with $\gamma \in [0, 1]$ as the discount factor, and $T$ as the horizon. The gradient of the expected reward $\triangledown_{\theta} J(\pi_\theta)$ can be determined according to the policy gradient theorem \citep{sutton2001policy}, which provides a direction of improvement to adjust the policy parameters $\theta$.

\[ \triangledown_{\theta} J(\pi_\theta) = \int_\mathcal{S} d_\theta(s) \int_\mathcal{A} \triangledown_{\theta} \mathrm{log}(\pi_\theta(s, a)) A(s, a) da \ ds \]

where $d_\theta(s) = \int_\mathcal{S} \sum_{t=0}^T \gamma^t p_0(s_0) p(s_0 \rightarrow s | t, \pi_\theta) ds_0$ is the discounted state distribution, where $p_0(s)$ represents the initial state distribution, and $p(s_0 \rightarrow s | t, \pi_\theta)$ models the likelihood of reaching state $s$ by starting at $s_0$ and following the policy $\pi_\theta(s, a)$ for $t$ steps \citep{icml2014c1_silver14}. $A(s, a)$ represents a generalized advantage function. The choice of advantage function gives rise to a family of policy gradient algorithms, but in this work, we will focus on the one-step temporal difference advantage function \citep{DBLP:journals/corr/SchulmanMLJA15}
\[ A(s_t, a_t) = r_t + \gamma V(s'_t) - V(s_t) \]
where $V(s) = \mathop{\mathbb{E}}\left[\sum_{t=0}^T \gamma^t r_t \middle| s_0 = s, \pi_\theta \right]$ is the state-value function, and can be defined recursively via the Bellman equation
\[ V(s_t) = \mathop{\mathbb{E}}_{r_t, s'_t} \left[ r_t + \gamma V(s'_t) \middle| s_t, \pi_\theta \right] \]

A parameterized value function $V_\phi(s)$, with parameters $\phi$, can be learned iteratively in a manner similar to Q-Learning by minimizing the Bellman loss,
\[ L(\phi) = \mathop{\mathbb{E}}_{s_t, r_t, s'_t} \left[ \frac{1}{2}\left( y_t - V_\phi(s_t) \right)^2 \right] , \qquad y_t = r_t + \gamma V_\phi(s'_t) \]

$\pi_\theta$ and $V_\phi$ can be trained in tandem using an actor-critic framework \citep{Konda00actor-criticalgorithms}.

In this work, each policy will be represented as a gaussian distribution with a parameterized mean $\mu_\theta(s)$ and fixed covariance matrix $\Sigma = \mathrm{diag}\{ \sigma_i^2 \}$, where $\sigma_i$ is manually specified for each action parameter. Actions can be sampled from the distribution by applying gaussian noise to the mean action
\[ a_t = \mu_\theta(s_t) + \mathcal{N}(0, \Sigma) \]
The corresponding policy gradient will assume the form
\[ \triangledown_{\theta} J(\pi_\theta) = \int_\mathcal{S} d_\theta(s) \int_\mathcal{A} \triangledown_{\theta} \mu_\theta(s) \Sigma^{-1} \left(a - \mu_\theta(s) \right) A(s, a) da \ ds \]

which can be interpreted as shifting the mean of the action distribution towards actions that lead to higher than expected rewards, while moving away from actions that lead to lower than expected rewards.

\section{Task Representation}

\subsection{Reference Motion}
In our task, the goal of a policy is to imitate a given reference motion $\{q^*_t\}$ which consists
of a sequence of kinematic poses $q^*_t$ in reduced coordinates. The reference velocity
$\dot{q^*_t}$ at a given time $t$ is approximated by finite-difference $ \dot{q^*_t} \approx
\frac{q^*_{t + \triangle t} - q^*_t}{\triangle t} $.  Reference motions are generated via either
using a recorded simulation result from a preexisting controller (``Sim''), or via manually-authored
keyframes. Since hand-crafted reference motions may not be physically realizable, the goal is to closely reproduce a
motion while satisfying physical constraints.

\subsection{States}
\begin{figure}[tbh]
\begin{centering}
\subfigure{   \includegraphics[width=0.33\columnwidth]{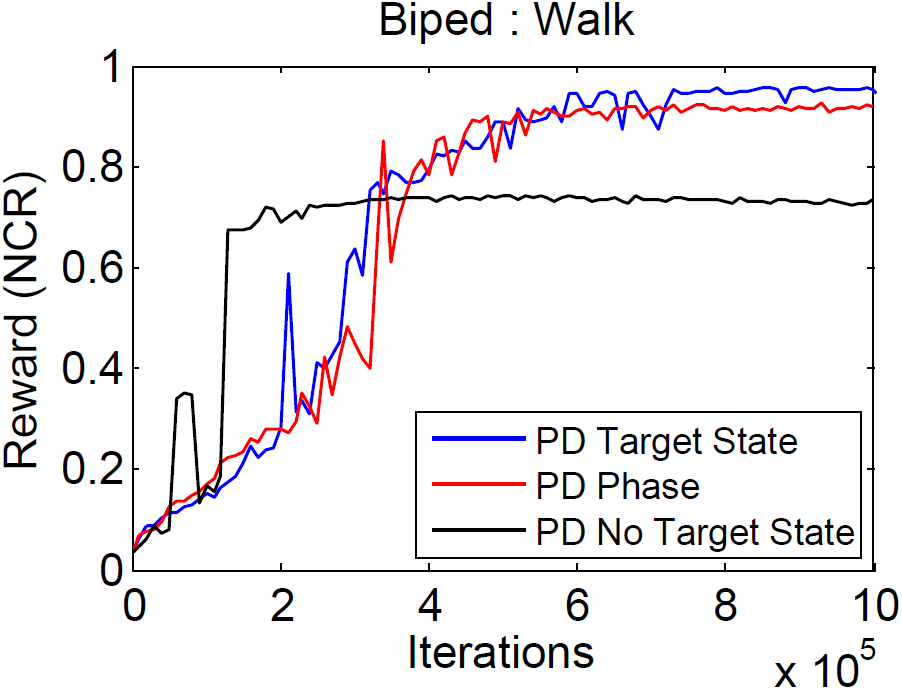}}
\subfigure{   \includegraphics[width=0.33\columnwidth]{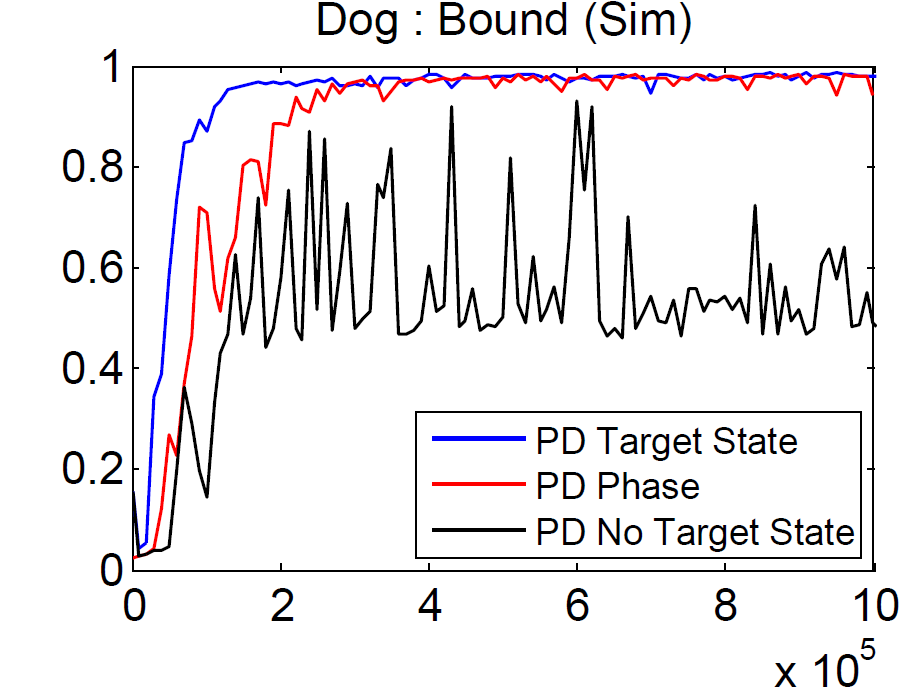}}
   \caption{Learning curves for different state representations including state + target state, state + phase, and only state.
     \label{fig:inputCurves}   }
\end{centering}
\end{figure}

To define the state of the agent, a feature transformation $\Phi(q, \dot{q})$ is used to extract
a set of features from the reduced-coordinate pose $q$ and velocity $\dot{q}$. The features consist
of the height of the root (pelvis) from the ground, the position of each link with respect to the
root, and the center of mass velocity of each link. When training a policy to imitate a cyclic reference
motion $\{q_t^*\}$, knowledge of the motion phase can help simplify learning. 
Therefore, we augment the state features with a set of target features $\Phi(q_t^*,
\dot{q_t}^*)$, resulting in a combined state represented by $s_t = (\Phi(q_t, \dot{q_t}),
\Phi(q_t^*, \dot{q_t}^*))$.  Similar results can also be achieved by providing a single
motion phase variable as a state feature, as we show in Figure~\ref{fig:inputCurves} (supplemental material).

\subsection{Actions}
We train separate policies for each of the four actuation models, as described below.
Each actuation model also has related actuation parameters, 
such as feedback gains for PD-controllers and musculotendon properties for MTUs. These
parameters can be manually specified, as we do for the PD and Vel models, or they can be optimized for the task at hand, 
as for the MTU models. Table~\ref{tab:actuationParams} provides a list of actuator parameters for each actuation model.

{\bf Target Joint Angles (PD):} Each action represents a set of target angles $\hat{q}$, where
$\hat{q}^i$ specifies the target angles for joint $i$. $\hat{q}$ is applied to PD-controllers
which compute torques according to
$ \tau^i = k_p^i (\hat{q}^i - q^i) + k^i_d(\hat{\dot{q}}^i - \dot{q}^i)$,
where $\hat{\dot{q}}^i = 0$, and $k_p^i$ and $k_d^i$ are manually-specified gains.

{\bf Target Joint Velocities (Vel):} Each action specifies a set of target velocities $\hat{\dot{q}}$
which are used to compute torques according to 
$ \tau^i = k_d^i(\hat{\dot{q}}^i - \dot{q}^i)$, where the gains $k_d^i$ are specified to be the same as those used for target angles.

{\bf Torques (Tor):}
Each action directly specifies torques for every joint, and constant torques are applied for the duration of a control step. Due to torque limits, actions are bounded by manually specified limits for each joint. Unlike the other actuation models, the torque model does not require additional actuator parameters, and can thus be regarded as requiring the least amount of domain knowledge. Torque limits are excluded from the actuator parameter set as they are common for all parameterizations.

{\bf Muscle Activations (MTU):}
Each action specifies activations for a set of musculotendon units (MTU). Detailed modeling and
implementation information are available in \citet{Wang12optimizinglocomotion}. Each MTU is modeled
as a contractile element (CE) attached to a serial elastic element (SE) and parallel elastic element
(PE). The force exerted by the MTU can be calculated according to $F_{MTU} = F_{SE} = F_{CE} +
F_{PE}$. Both $F_{SE}$ and $F_{PE}$ are modeled as passive springs, while $F_{CE}$ is actively
controlled according to $ F_{CE} = a_{MTU} F_0 f_l(l_{CE}) f_v(v_{CE})$, with $a_{MTU}$ being the
muscle activation, $F_0$ the maximum isometric force, $l_{CE}$ and $v_{CE}$ being the length and
velocity of the contractile element. The functions $f_l(l_{CE})$ and $f_v(v_{CE})$ represent the
force-length and force-velocity relationships, modeling the variations in the maximum force that can
be exerted by a muscle as a function of its length and contraction velocity. Analytic forms are
available in \citet{Geyer2173}. Activations are bounded between [0, 1]. The length of each
contractile element $l_{CE}$ are included as state features. To simplify control and reduce the
number of internal state parameters per MTU, the policies directly control muscle activations
instead of indirectly through excitations \citep{Wang12optimizinglocomotion}. 



\begin{table}[tbh]
{ \centering  
\begin{tabular}{|c|c|}
\hline
{\bf Actuation Model} & {\bf Actuator Parameters} \\ \hline
Target Joint Angles (PD) & proportional gains $k_p$, derivative gains $k_d$\\ \hline
Target Joint Velocities (Vel) & derivative gains $k_d$ \\ \hline
Torques (Tor) & none \\ \hline
Muscle Activations (MTU) & optimal contractile element length, serial elastic element rest length, \\
& maximum isometric force, pennation, moment arm, \\
& maximum moment arm joint orientation, rest joint orientation. \\ \hline
\end{tabular} \\
}
\caption{Actuation models and their respective actuator parameters.}
\label{tab:actuationParams}
\end{table}

\subsection{Reward}
The reward function consists of a weighted sum of terms that encourage the policy to track a reference motion.
\[ r = w_{pose} r_{pose} + w_{vel} r_{vel} + w_{end} r_{end} + w_{root} r_{root} + w_{com} r_{com} \]
\[w_{pose} = 0.5, \ w_{vel} = 0.05, \ w_{end} = 0.15, \ w_{root} = 0.1, \ w_{com} = 0.2\]

Details of each term are available in the supplemental material. $r_{pose}$ penalizes deviation of the character pose from the reference pose, and $r_{vel}$ penalizes deviation of the joint velocities. $r_{end}$  and $r_{root}$ accounts for the position error of the end-effectors and root. $r_{com}$ penalizes deviations in the center of mass velocity from that of the reference motion.

\subsection{Initial State Distribution}


We design the initial state distribution, $p_0(s)$, to sample states uniformly along the reference trajectory. 
At the start of each episode, $q^*$ and $\dot{q}^*$ are sampled from the reference trajectory, and used to initialize the pose and
velocity of the agent. This helps guide the agent to explore states near the target trajectory.

\section{Actor-Critic Learning Algorithm}
Instead of directly using the temporal difference advantage function, we adapt a positive temporal
difference (PTD) update as proposed by \citet{vanHasselt2012}.
\[ A(s, a) = I\left[ \delta > 0 \right] = \begin{cases}
    1, 		& \text{$\delta > 0$} \\
    0,      & \text{otherwise}
\end{cases} \]
\[ \delta = r + \gamma V(s') - V(s) \]
Unlike more conventional policy gradient methods, PTD is less sensitive to the scale of the
advantage function and avoids instabilities that can result from negative TD updates. For a Gaussian
policy, a negative TD update moves the mean of the distribution away from an observed action,
effectively shifting the mean towards an unknown action that may be no better than the current mean
action~\citep{vanHasselt2012}. In expectation, these updates converges to the true policy gradient,
but for stochastic estimates of the policy gradient, these updates can cause the agent to adopt
undesirable behaviours which affect subsequent experiences collected by the agent. Furthermore, we
incorporate experience replay, which has been demonstrated to improve stability when training neural
network policies with Q-learning in discrete action spaces. Experience replay
often requires off-policy methods, such as importance weighting, to account for differences between
the policy being trained and the behavior policy used to generate experiences~\citep{wawrzynski2013}. 
However, we have not found importance weighting to be beneficial for PTD.

Stochastic policies are used during training for exploration, while deterministic policy are
deployed for evaluation at runtime. The choice between a stochastic and deterministic policy can be
specified by the addition of a binary indicator variable $\lambda \in [0, 1]$
\[ a_t = \mu_\theta(s_t) + \lambda \mathcal{N}(0, \Sigma) \]
where $\lambda = 1$ corresponds to a stochastic policy with exploration noise, and $\lambda = 0$
corresponds to a deterministic policy that always selects the mean of the distribution. Noise from a
stochastic policy will result in a state distribution that differs from that of the deterministic
policy at runtime. To imitate this discrepancy, we incorporate $\epsilon$-greedy exploration in
addition to the original Gaussian exploration. During training, $\lambda$ is determined by a
Bernoulli random variable $\lambda \sim \mathrm{Ber}(\epsilon)$, where $\lambda = 1$ with
probability $\epsilon \in [0, 1]$. The exploration rate $\epsilon$ is annealed linearly from 1 to
0.2 over 500k iterations, which slowly adjusts the state distribution encountered during training to
better resemble the distribution at runtime. Since the policy gradient is defined for stochastic
policies, only tuples recorded with exploration noise (i.e. $\lambda = 1$) can be used to update the
actor, while the critic can be updated using all tuples.

Training proceeds episodically, where the
initial state of each episode is sampled from $p_0(s)$, and the episode duration is drawn from an
exponential distribution with a mean of 2s. To discourage falling, an episode will also terminate if
any part of the character's trunk makes contact with the ground for an extended period of time,
leaving the agent with zero reward for all subsequent steps.
Algorithm \ref{alg:CACLA} in the supplemental material summarizes the complete learning process.

{\bf MTU Actuator Optimization:}
Actuation models such as MTUs are defined by further parameters whose values impact performance~\citep{2013-TOG-MuscleBasedBipeds}. 
\citet{Geyer2173} uses existing anatomical estimates for humans to determine MTU parameters, but such data is not be available for more arbitrary
creatures. Alternatively, \citet{2013-TOG-MuscleBasedBipeds} uses covariance
matrix adaptation (CMA), a derivative-free evolutionary search strategy, to simultaneously optimize
MTU and policy parameters. This approach is limited to policies with reasonably low
dimensional parameter spaces, and is thus ill-suited for neural network models with hundreds of
thousands of parameters. To avoid manual-tuning of actuator parameters, we
propose a heuristic approach that alternates between policy learning and actuator optimization,
as detailed in the supplemental material.


\begin{figure}[tbh]
\begin{centering}
\subfigure{   \includegraphics[width=0.2\columnwidth]{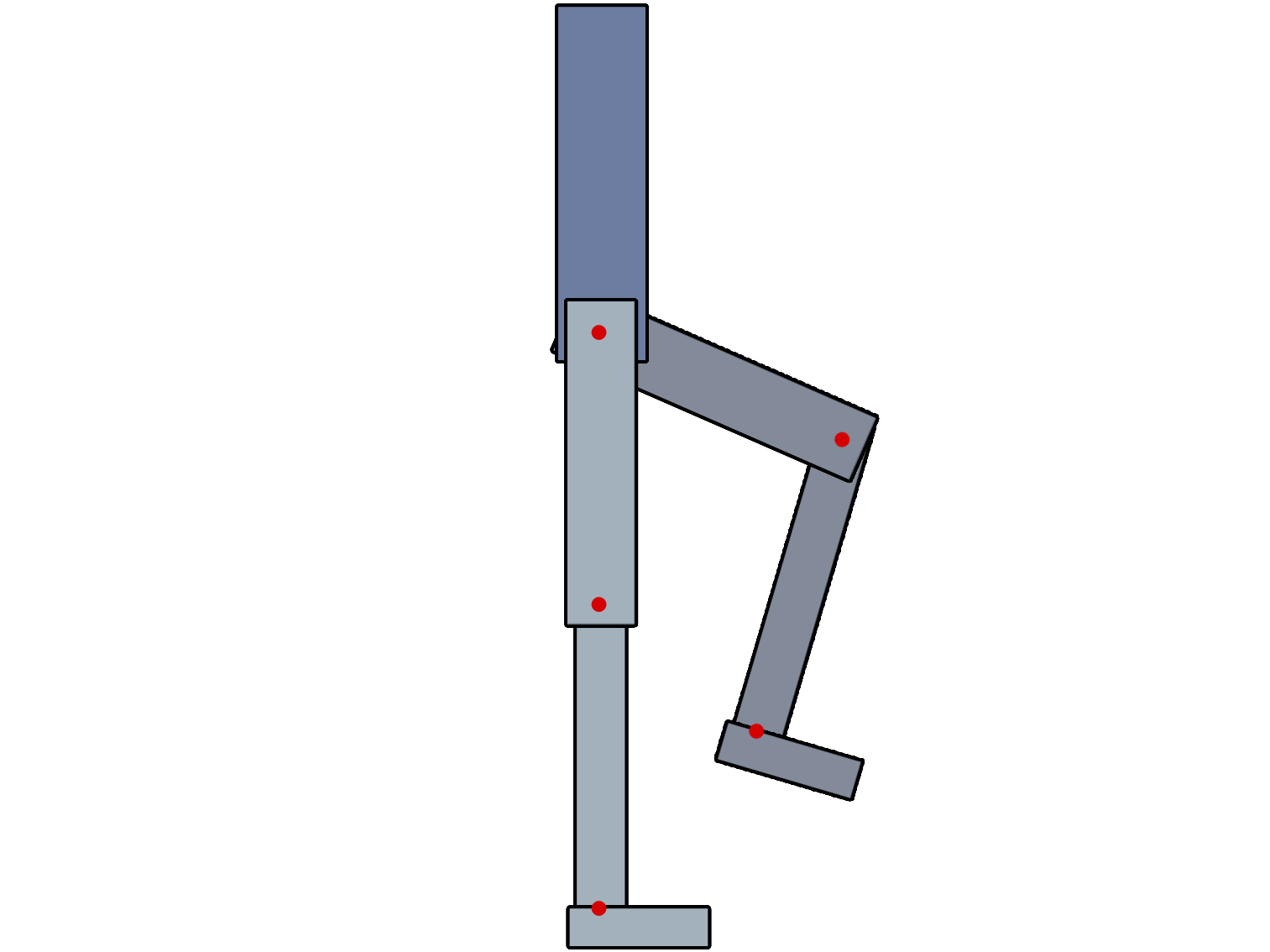}}
\subfigure{   \includegraphics[width=0.25\columnwidth]{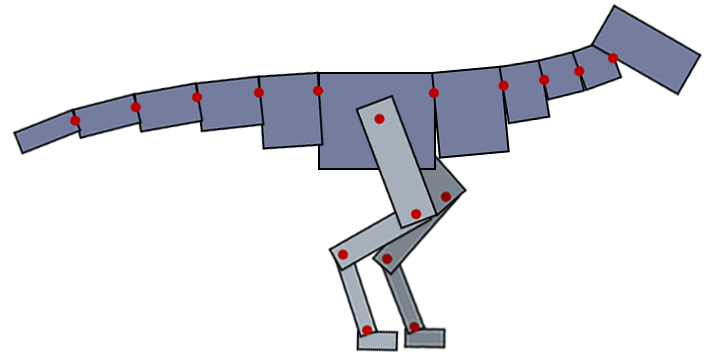}}
\hspace*{0.25cm} \subfigure{   \includegraphics[width=0.18\columnwidth]{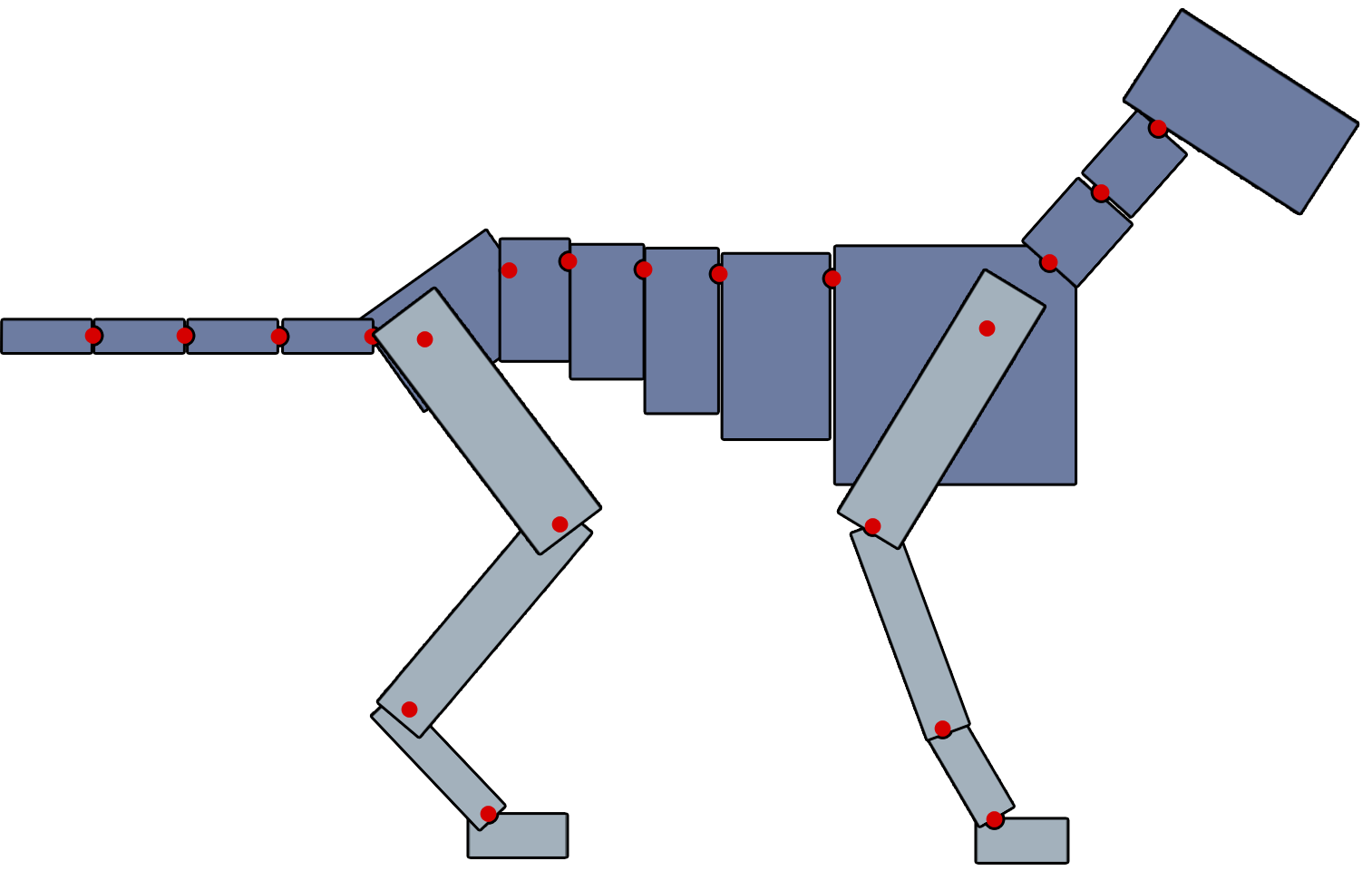}}
\hspace*{0.4cm} 
\subfigure{   \includegraphics[width=0.2\columnwidth]{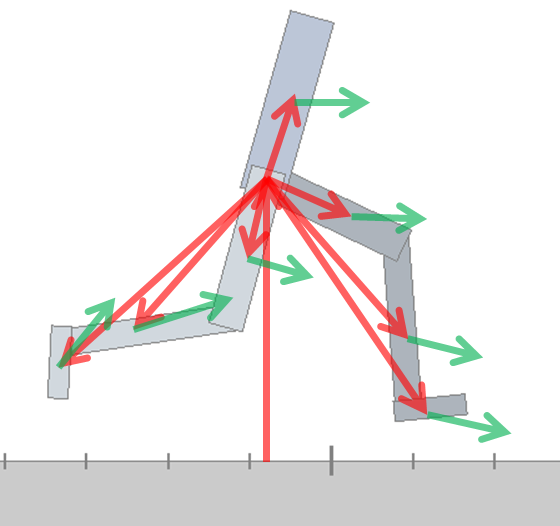}}
   \caption{Simulated articulated figures and their state representation. Revolute joints connect all links. From left to right: 
   7-link biped; 19-link raptor; 21-link dog; State features: root height, relative position (red) of each link with respect to the root and their respective linear velocity (green).
     \label{fig:characters}   }
\end{centering}
\end{figure}


\begin{figure}[t!]
\begin{centering}
\subfigure{   \includegraphics[width=0.245\columnwidth]{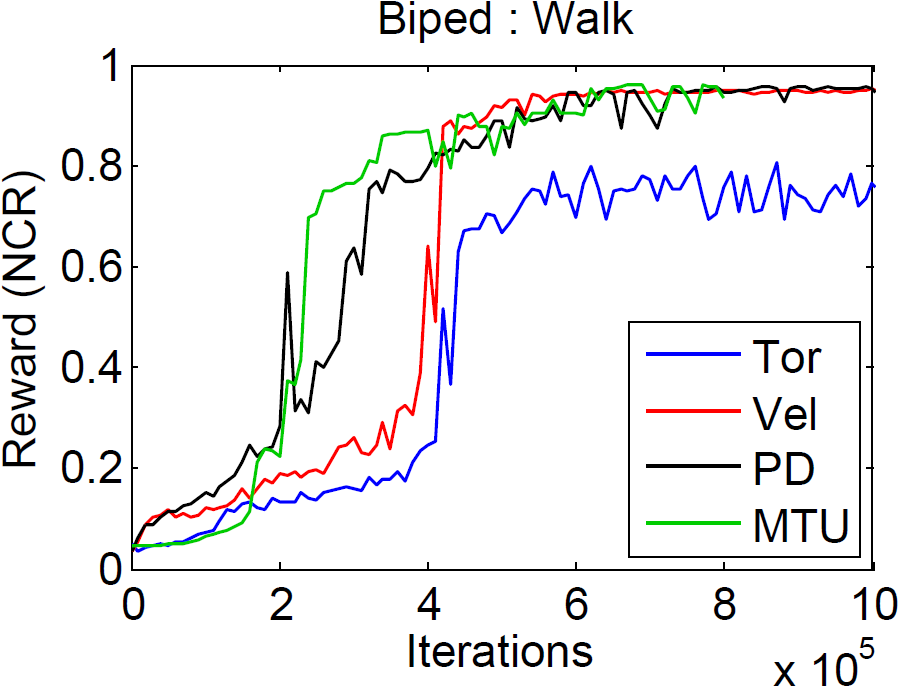}}
\subfigure{   \includegraphics[width=0.23\columnwidth]{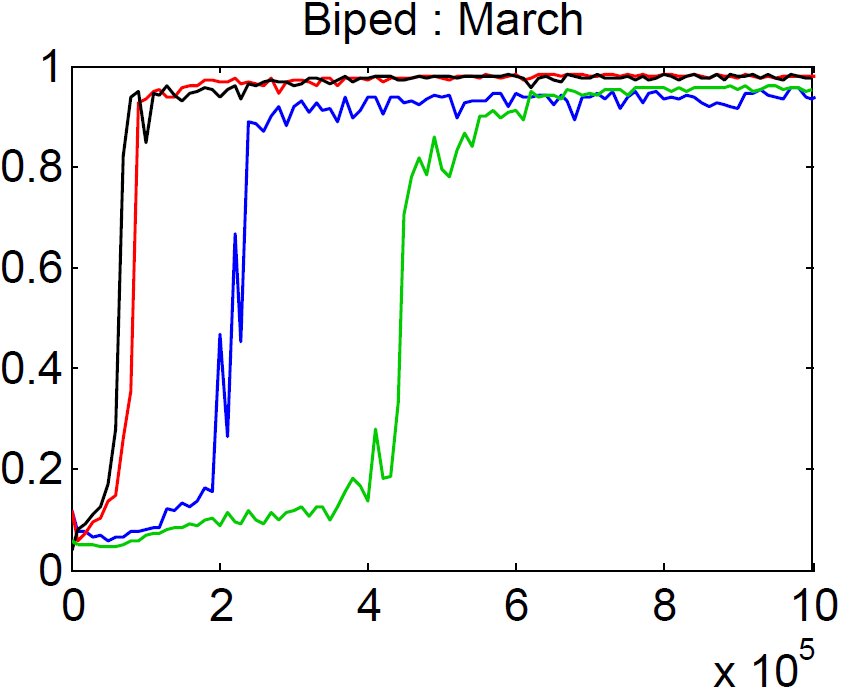}}
\subfigure{   \includegraphics[width=0.23\columnwidth]{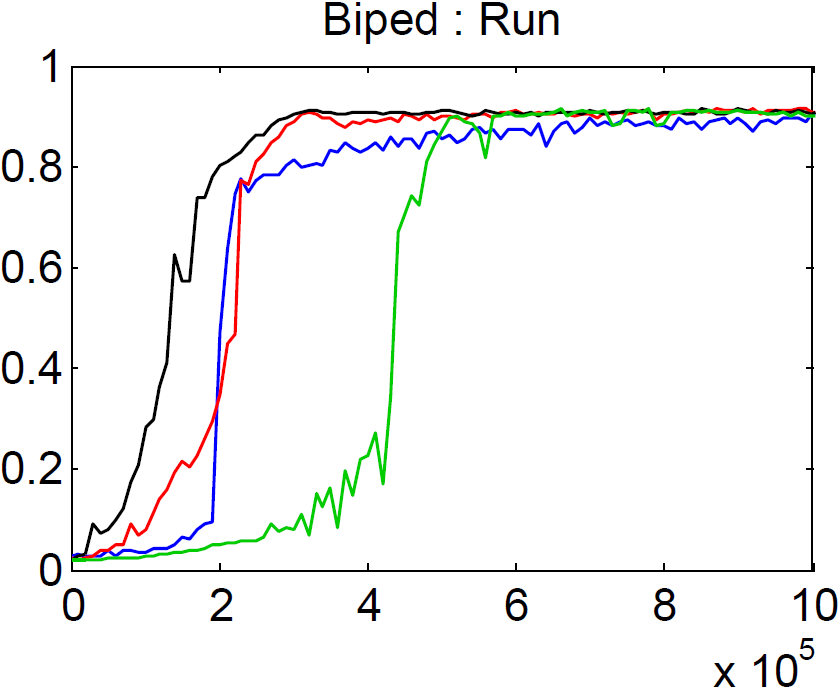}}
\subfigure{   \includegraphics[width=0.23\columnwidth]{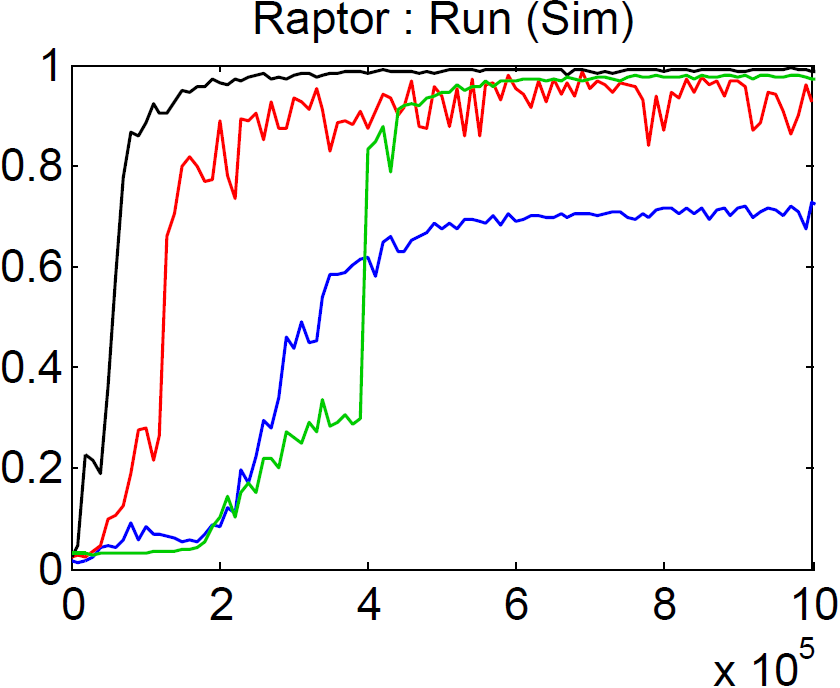}}
\subfigure{   \includegraphics[width=0.23\columnwidth]{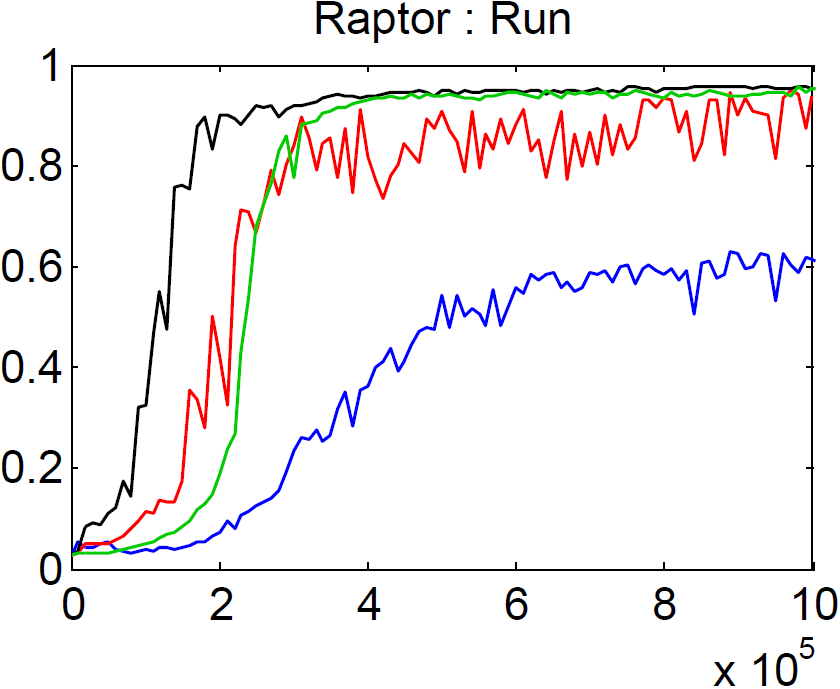}}
\subfigure{   \includegraphics[width=0.23\columnwidth]{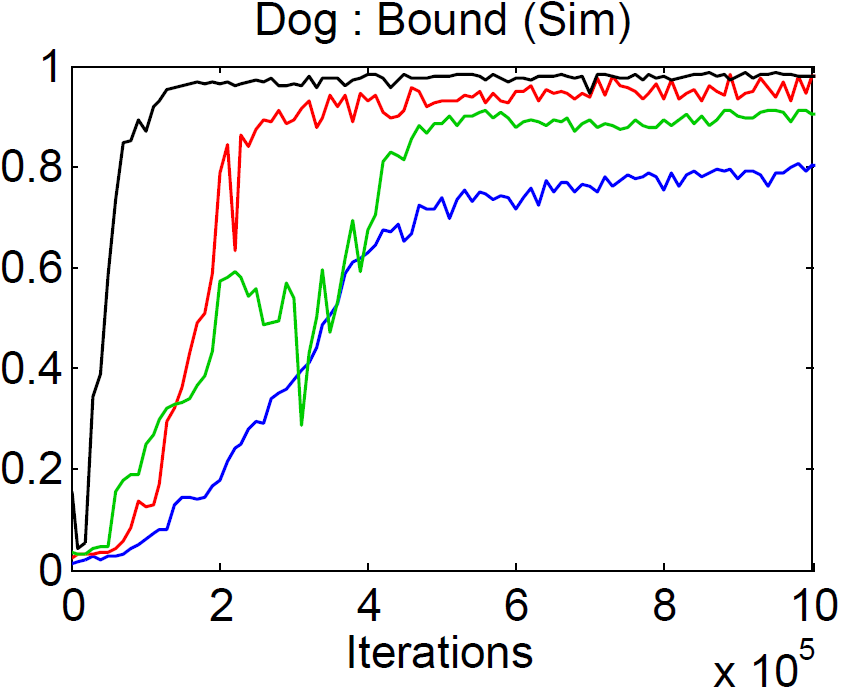}}
\subfigure{   \includegraphics[width=0.23\columnwidth]{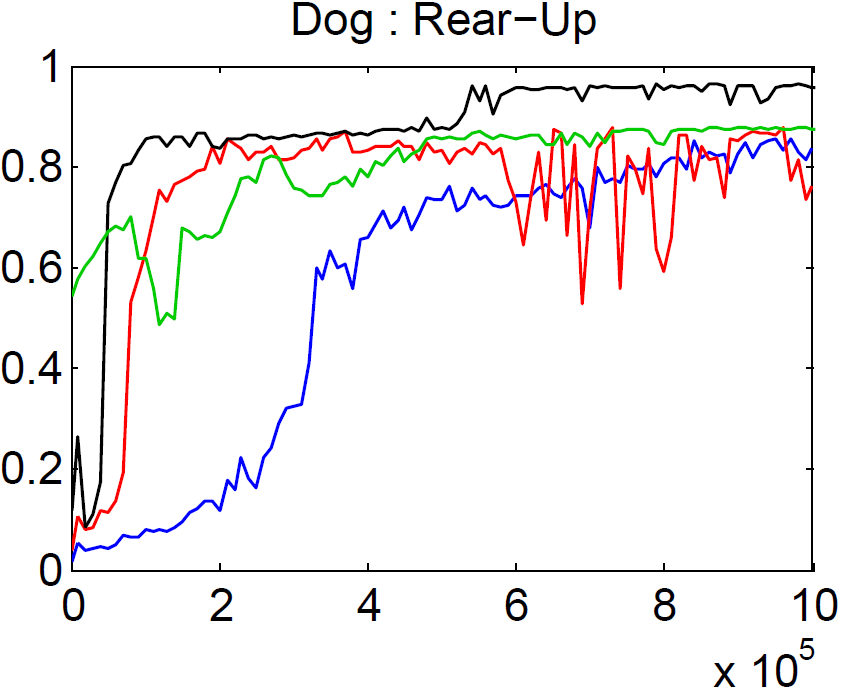}}
   \caption{Learning curves for each policy during 1 million iterations.
     \label{fig:learningCurves}   }
\end{centering}
\end{figure}

\section{Results}

The motions are best seen in the supplemental video https://youtu.be/L3vDo3nLI98. We evaluate the action
parameterizations by training policies for a simulated 2D biped, dog, and raptor as shown in Figure~\ref{fig:characters}.
Depending on the agent and the actuation model, our systems have 58--214 state dimensions, 
6--44 action dimensions, and 0--282 actuator parameters,
as summarized in Table~\ref{tab:spaceDims} (supplemental materials). The MTU models have at least double
the number of action parameters because they come in antagonistic pairs. As well, additional MTUs
are used for the legs to more accurately reflect bipedal biomechanics. This includes MTUs that span
multiple joints.

\begin{figure}[tbh]
\begin{centering}
\subfigure{   \includegraphics[width=0.5\columnwidth]{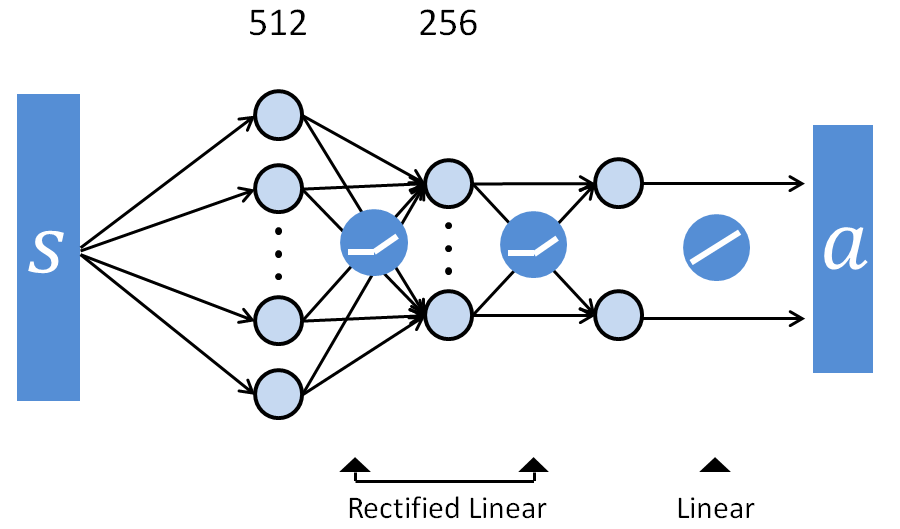}} 
   \caption{Neural Network Architecture. Each policy is represented by a three layered network, 
with 512 and 256 fully-connected hidden units, followed by a linear output layer.
     \label{fig:net}   }
\end{centering}
\end{figure}

Each policy is represented by a three layer neural network, as illustrated in Figure~\ref{fig:net} (supplemental material) 
with 512 and 256 fully-connected units, followed by a linear output layer where the number of output units vary according to the number of
action parameters for each character and actuation model. ReLU activation functions are used for
both hidden layers. Each network has approximately 200k parameters. 
The value function is represented by a similar network, except having a single linear output unit. The
policies are queried at 60Hz for a control step of about 0.0167s. Each network is randomly
initialized and trained for about 1 million iterations, requiring 32 million tuples, the equivalent
of approximately 6 days of simulated time. Each policy requires about 10 hours for the biped, and 20
hours for the raptor and dog on an 8-core Intel Xeon E5-2687W.


Only the actuator parameters for MTUs are optimized with Algorithm \ref{alg:mtuOpt}, since the parameters for the other actuation models are few and reasonably intuitive to determine. The initial actuator parameters $\psi_0$ are manually specified, while the initial policy parameters $\theta_0$ are randomly initialized. Each pass optimizes $\psi$ using CMA for 250 generations with 16 samples per generation, and $\theta$ is trained for 250k iterations. Parameters are initialized with values from the previous pass.
The expected value of each CMA sample of $\psi$ is estimated using the average cumulative reward over 16 rollouts with a duration of 10s each. Separate MTU parameters are optimized for each character and motion. Each set of parameters is optimized for 6 passes following Algorithm \ref{alg:mtuOpt}, requiring approximately 50 hours. Figure \ref{fig:optMTUCurves} illustrates the performance improvement per pass. Figure \ref{fig:optMTUCompare} compares the performance of MTUs before and after optimization. For most examples, the optimized actuator parameters significantly improve learning speed and final performance. For the sake of comparison, after a set of actuator parameters has been optimized, a new policy is retrained with the new actuator parameters and its performance compared to the other actuation models.

{\bf Policy Performance and Learning Speed:}
Figure~\ref{fig:learningCurves} shows learning curves for the policies and the performance of the final policies are summarized in 
Table~\ref{tab:perf}. Performance is evaluated using the normalized cumulative reward (NCR), calculated from the average cumulative reward over 32 episodes with lengths of 10s, and normalized by the maximum and minimum cumulative reward possible for each episode. No discounting is applied when calculating the NCR. The initial state of each episode is sampled from the reference motion according to $p(s_0)$. To compare learning speeds, we use the normalized area under each learning curve (AUC) as a proxy for the learning speed of a particular actuation model, where 0 represents the worst possible performance and no progress during training, and 1 represents the best possible performance without requiring training.

PD performs well across all examples, achieving comparable-to-the-best performance for all motions. PD also learns faster than the other parameterizations for 5 of the 7 motions. The final performance of Tor is among the poorest for all the motions. Differences in performance appear more pronounced as characters become more complex. For the simple 7-link biped, most parameterizations achieve similar performance. However, for the more complex dog and raptor, the performance of Tor policies deteriorate with respect to other policies such as PD and Vel. MTU policies often exhibited the slowest learning speed, which may be a consequence of the higher dimensional action spaces, i.e., requiring antagonistic muscle pairs, and complex muscle dynamics. Nonetheless, once optimized, the MTU policies produce more natural motions and responsive behaviors as compared to other parameterizations. We note that the naturalness of motions is not well captured by the reward, since it primarily gauges similarity to the reference motion, which may not be representative of natural responses when perturbed from the nominal trajectory.

\begin{figure}[bth]
\begin{centering}
\subfigure{   \includegraphics[width=0.32\columnwidth]{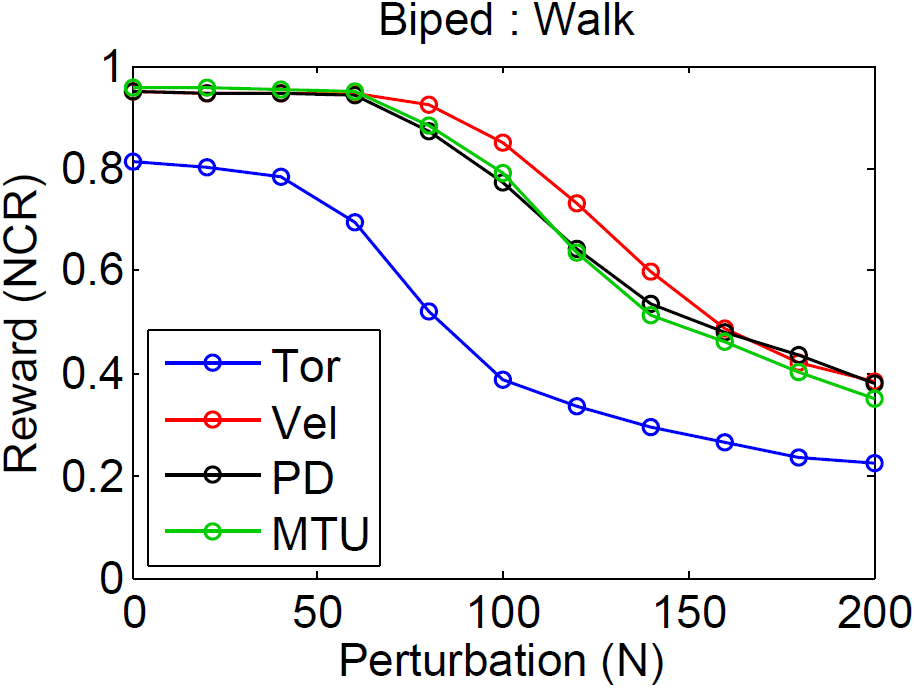}}
\subfigure{   \includegraphics[width=0.32\columnwidth]{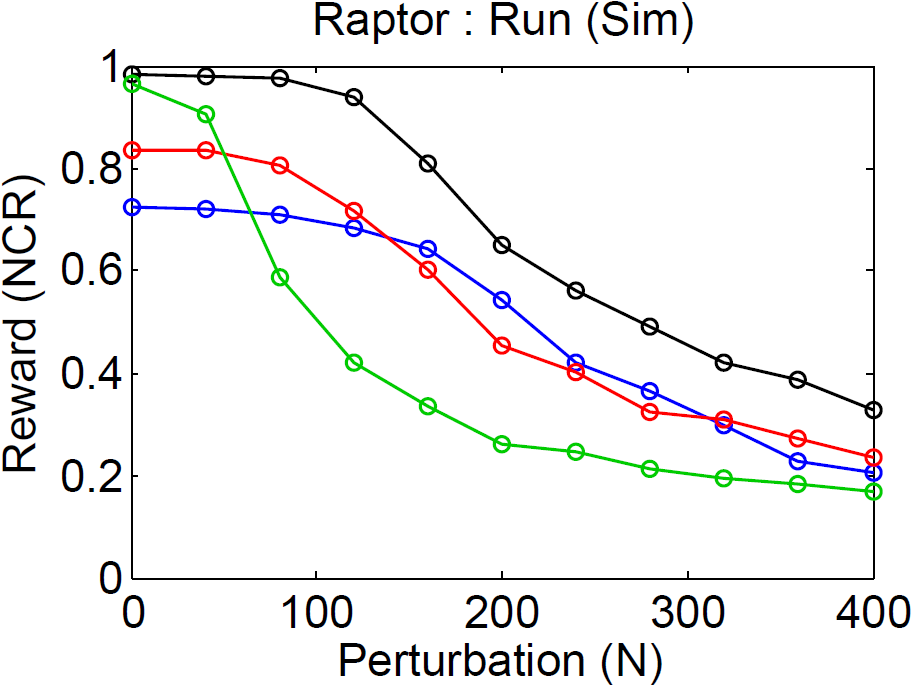}}
\subfigure{   \includegraphics[width=0.32\columnwidth]{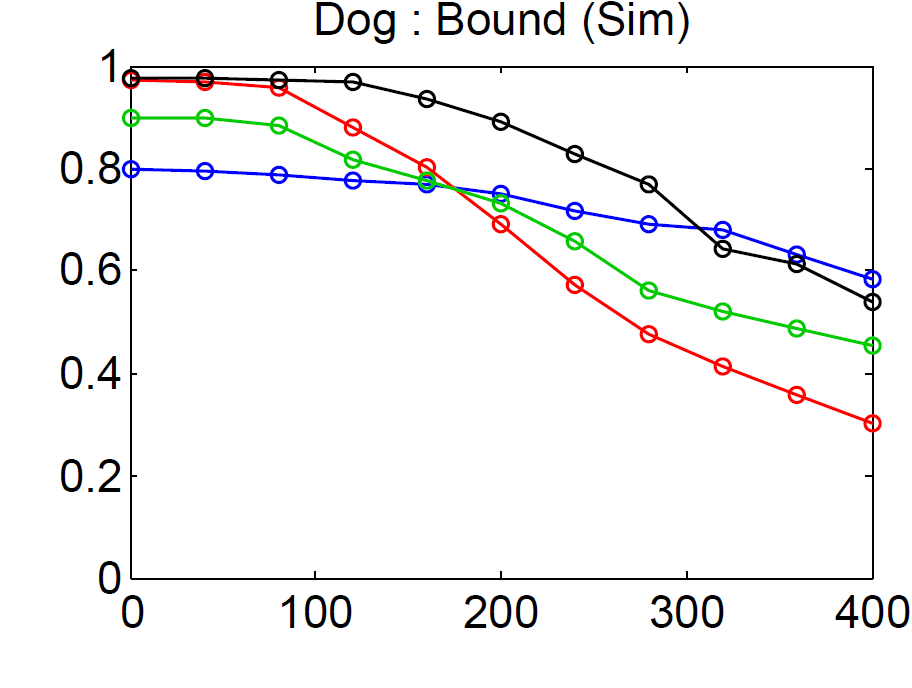}}
   \caption{Performance when subjected to random perturbation forces of different magnitudes.
     \label{fig:perturbPerf}   }
\end{centering}
\end{figure}

\begin{figure}[h]
\begin{centering}
\subfigure{   \includegraphics[width=0.32\columnwidth]{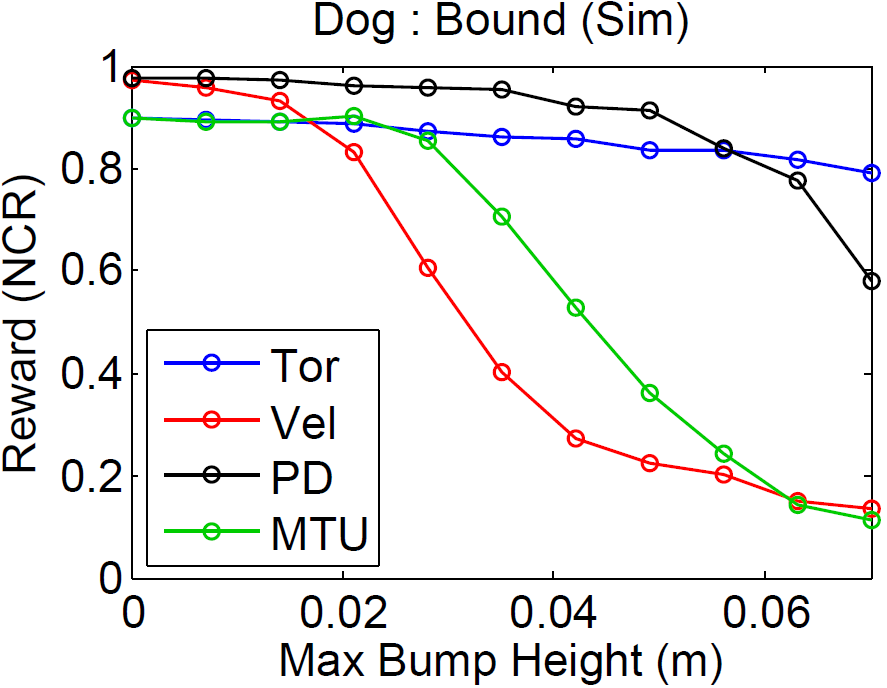}}
\subfigure{   \includegraphics[width=0.32\columnwidth]{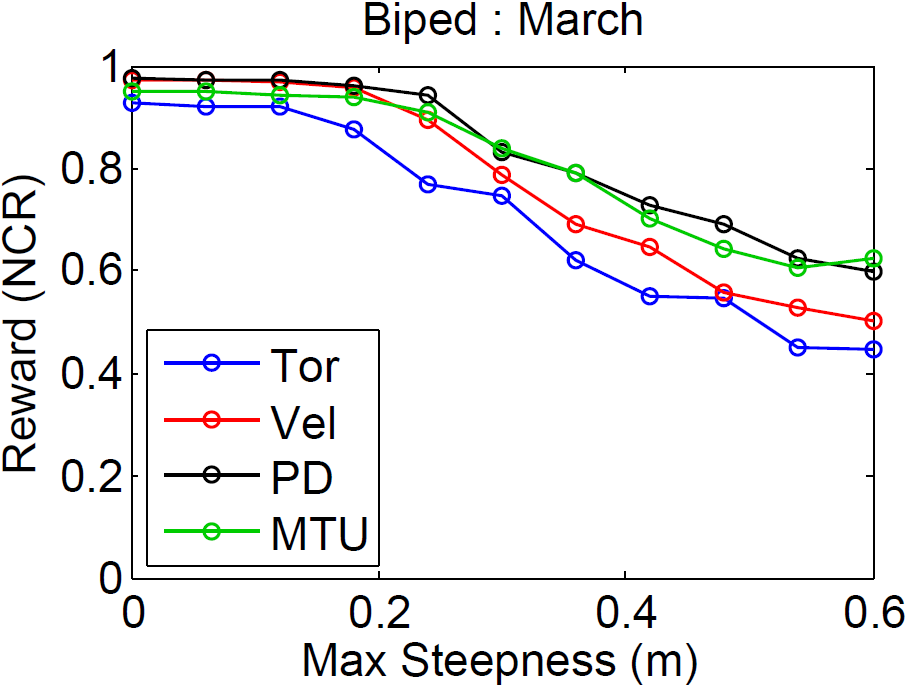}}
\subfigure{   \includegraphics[width=0.32\columnwidth]{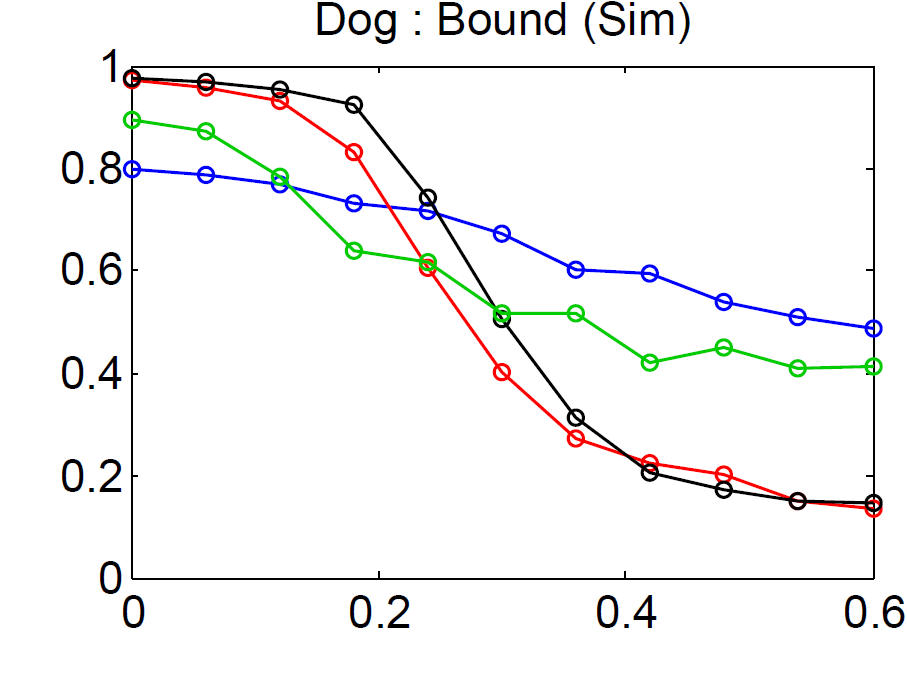}}
   \caption{Performance of different action parameterizations when traveling across randomly generated irregular terrain. \textbf{(left)} Dog running across bumpy terrain, where the height of each bump varies uniformly between 0 and a specified maximum height. \textbf{(middle)} and \textbf{(right)} biped and dog traveling across randomly generated slopes with bounded maximum steepness.
     \label{fig:terrainPerf}   }
\end{centering}
\end{figure}

{\bf Policy Robustness:} To evaluate robustness, we recorded the NCR achieved by each policy when subjected to external perturbations. The perturbations assume the form of random forces applied to the trunk of the characters. Figure~\ref{fig:perturbPerf} illustrates the performance of the policies when subjected to perturbations of different magnitudes. The magnitude of the forces are constant, but direction varies randomly. Each force is applied for 0.1 to 0.4s, with 1 to 4s between each perturbation. Performance is estimated using the average over 128 episodes of length 20s each. For the biped walk, the Tor policy is significantly less robust than those for the other types of actions, while the MTU policy is the least robust for the raptor run. Overall, the PD policies are among the most robust for all the motions.
In addition to external forces, we also evaluate robustness over randomly generated terrain consisting of bumps with varying heights and slopes with varying steepness. We evaluate the performance on irregular terrain (Figure~\ref{fig:terrainPerf}, supplemental material). There are few discernible patterns for this test. The Vel and MTU policies are significantly worse than the Tor and PD policies for the dog bound on the bumpy terrain. The unnatural jittery behavior of the dog Tor policy proves to be surprisingly robust for this scenario. We suspect that the behavior prevents the trunk from contacting the ground for extended periods for time, and thereby escaping our system's fall detection.



\begin{figure}[bth]
\begin{centering}
\subfigure{   \includegraphics[width=0.33\columnwidth]{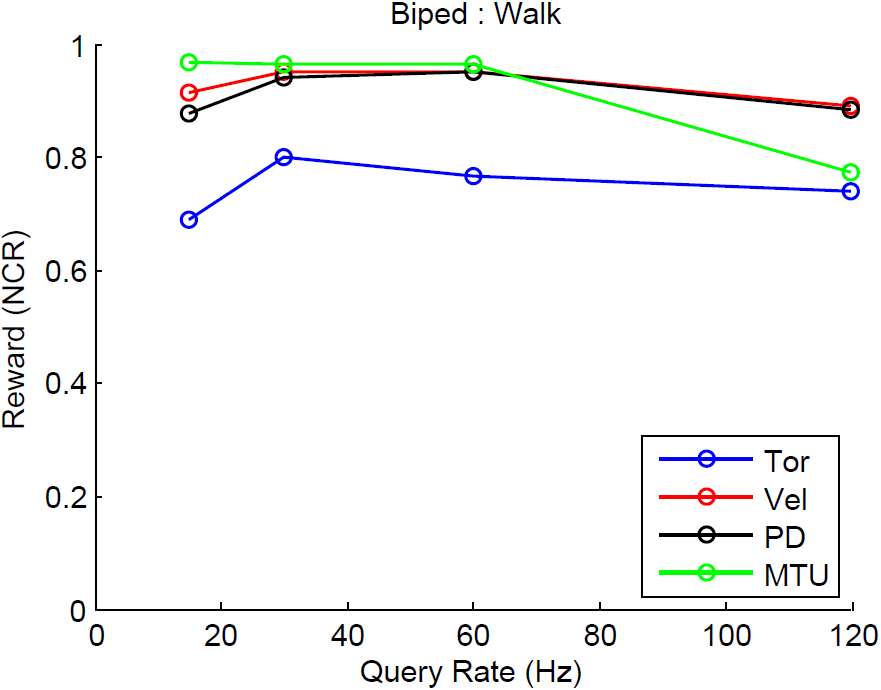}}
\subfigure{   \includegraphics[width=0.33\columnwidth]{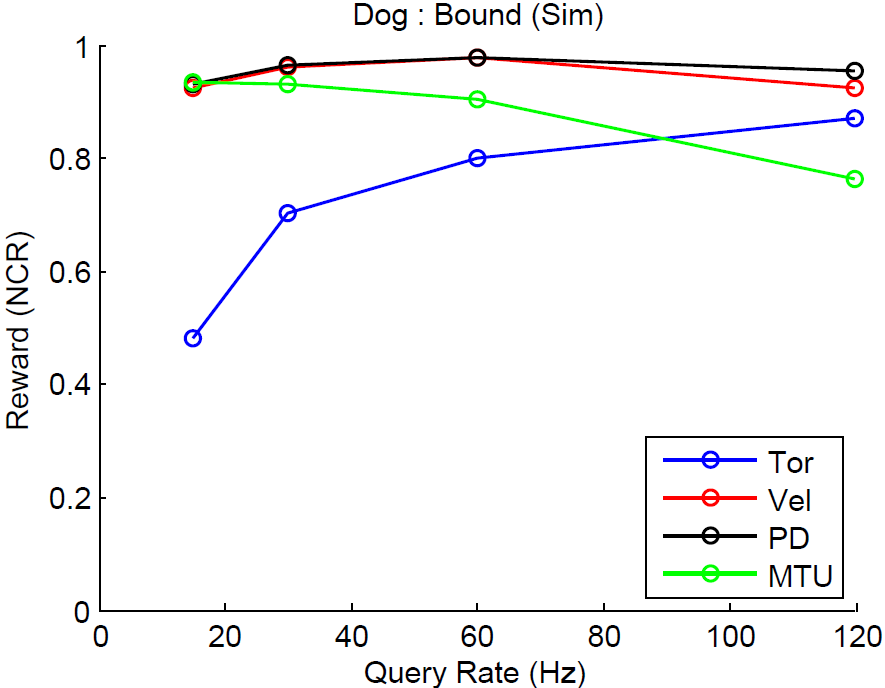}}
   \caption{Performance of policies with different query rates for the biped \textbf{(left)} and dog \textbf{(right)}. Separate policies are trained for each query rate.
     \label{fig:queryRate}   }
\end{centering}
\end{figure}

\begin{figure}[tbh]
\begin{centering}
\vspace*{-0.5cm}
\subfigure{   \includegraphics[width=0.43\columnwidth]{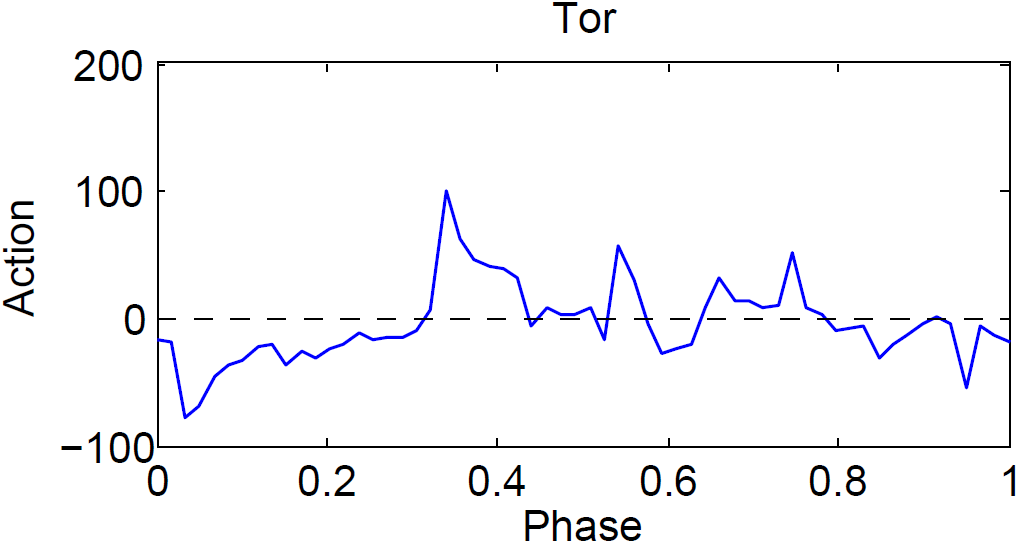}} \hspace{0.00\columnwidth}
\subfigure{   \includegraphics[width=0.43\columnwidth]{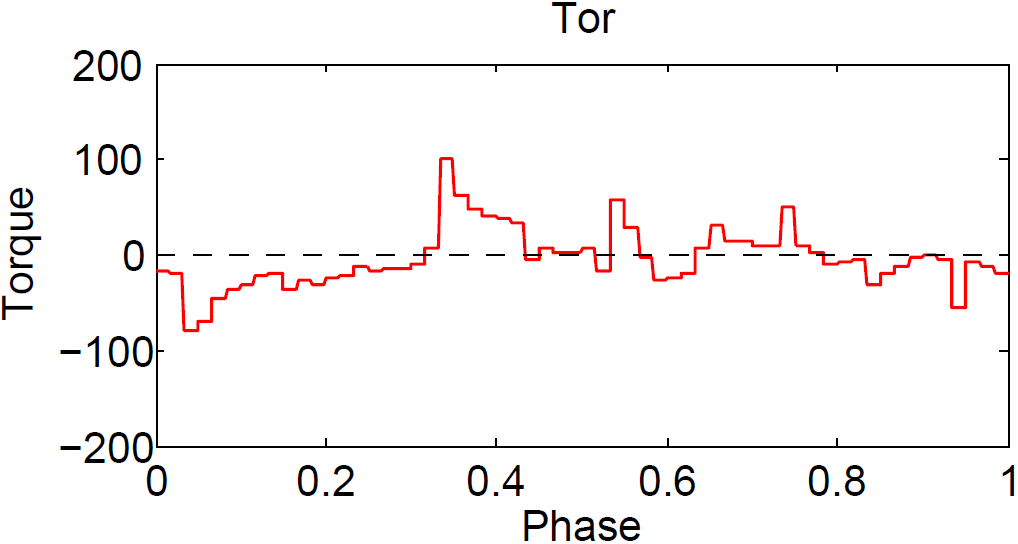}}
\subfigure{   \includegraphics[width=0.43\columnwidth]{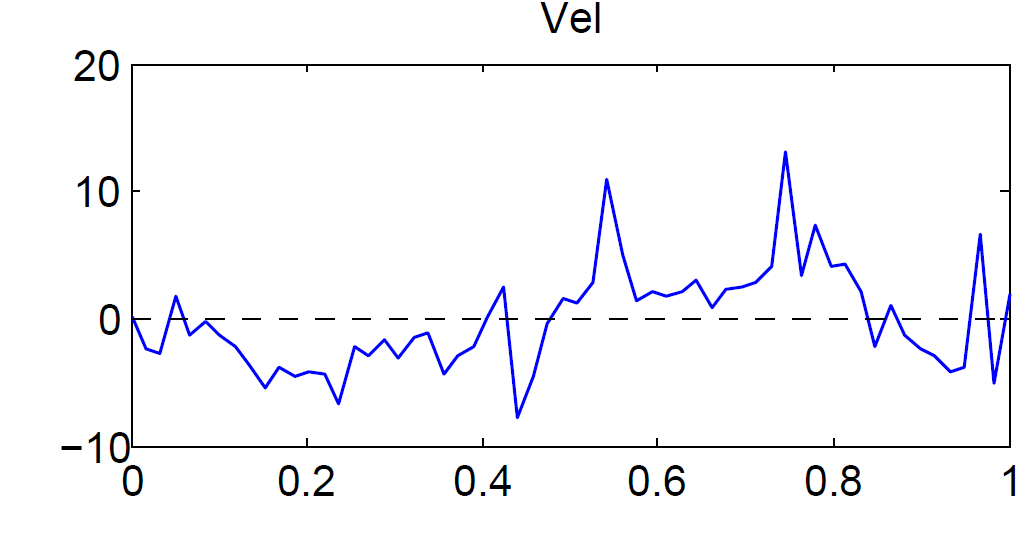}} \hspace{0.00\columnwidth}
\subfigure{   \includegraphics[width=0.43\columnwidth]{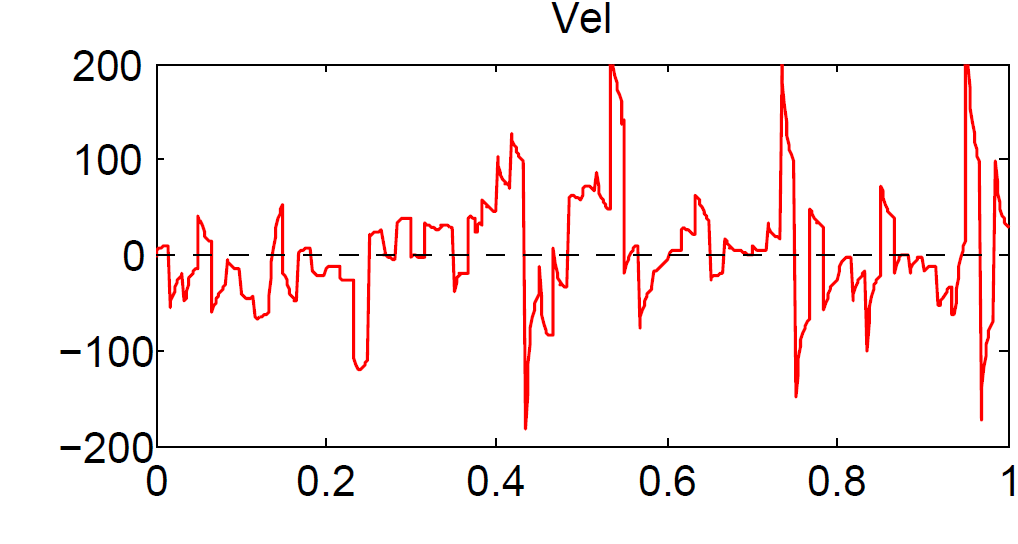}}
\subfigure{   \includegraphics[width=0.43\columnwidth]{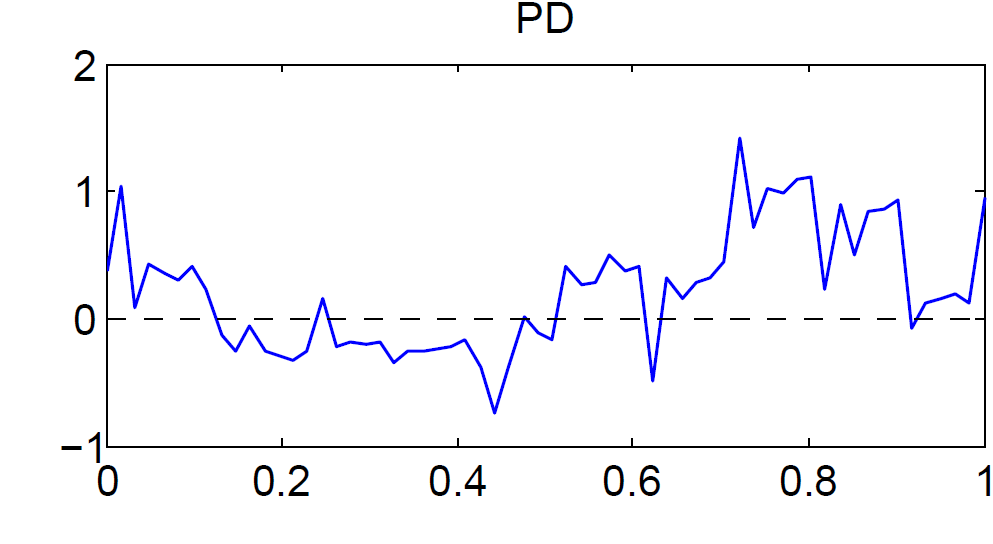}} \hspace{0.00\columnwidth}
\subfigure{   \includegraphics[width=0.43\columnwidth]{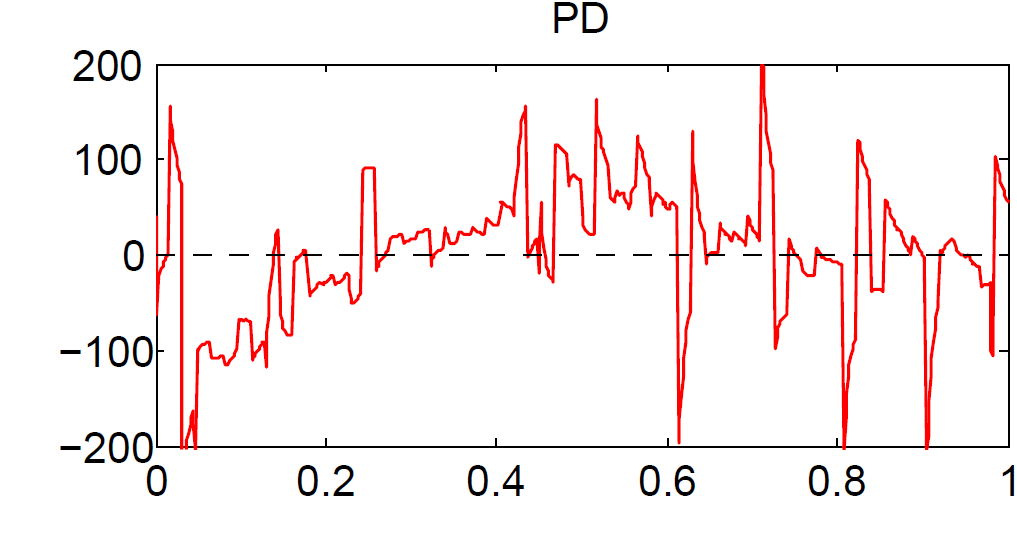}}
\subfigure{   \includegraphics[width=0.43\columnwidth]{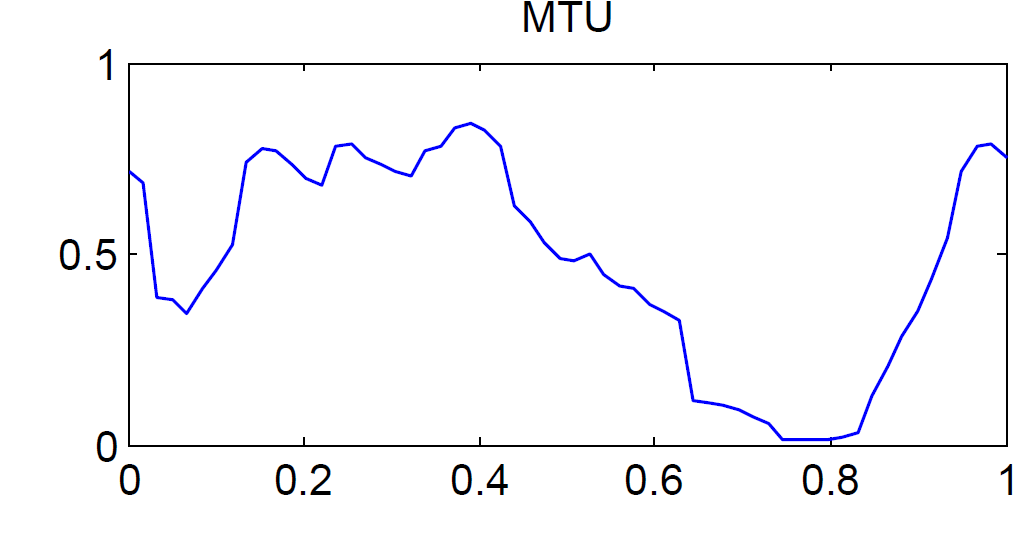}} \hspace{0.00\columnwidth}
\subfigure{   \includegraphics[width=0.43\columnwidth]{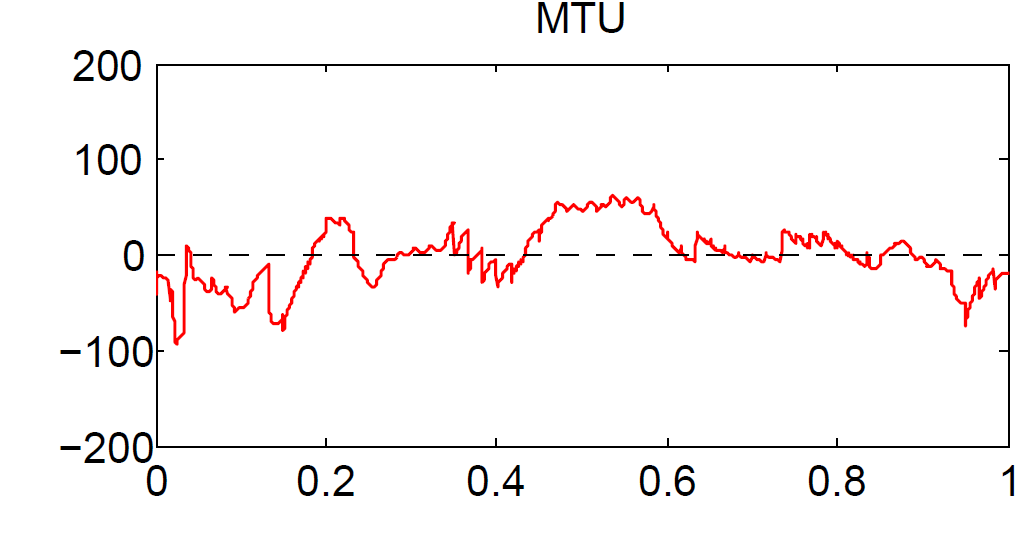}}
\hspace{0.00\columnwidth}
   \caption{Policy actions over time and the resulting torques for the four action types. Data is from one biped walk cycle (1s). 
   Left: Actions (60~Hz), for the right hip for PD, Vel, and Tor, and the right gluteal muscle for MTU. 
   Right: Torques applied to the right hip joint, sampled at 600~Hz.
     \label{fig:trajectories}   }
\end{centering}
\end{figure}

{\bf Query Rate:}  Figure~\ref{fig:queryRate} compares the performance of different parameterizations for different policy query rates. Separate policies are trained with queries of 15Hz, 30Hz, 60Hz, and 120Hz. Actuation models that incorporate low-level feedback such as PD and Vel, appear to cope more effectively to lower query rates, while the Tor degrades more rapidly at lower query rates. It is not yet obvious to us why MTU policies appear to perform better at lower query rates and worse at higher rates. 
Lastly, Figure~\ref{fig:trajectories} shows the policy outputs as a function of time for the four actuation models, for a particular joint, as well as showing the resulting joint torque. Interestingly, the MTU action is visibly smoother than the other actions and results in joint torques profiles that are smoother than those seen for PD and Vel. 



\section{Related Work}

DeepRL has driven impressive recent advances in learning motion control, i.e., solving for continuous-action control problems  using reinforcement learning. All four of the actions types that we explore have seen previous use in the machine learning literature.  \citet{wawrzynski2013} use an actor-critic approach with experience replay to learn skills for an octopus arm (actuated by a simple muscle model) and a planar half cheetah (actuated by joint-based PD-controllers). Recent work on deterministic policy gradients~\citep{DBLP:journals/corr/LillicrapHPHETS15} and on RL benchmarks, e.g., OpenAI Gym, generally use joint torques as the action space, as do the test suites in recent work ~\citep{DBLP:journals/corr/SchulmanMLJA15} on using generalized advantage estimation. Other recent work uses: the PR2 effort control interface as a proxy for torque control~\citep{DBLP:journals/corr/LevineFDA15}; joint velocities~\citep{gu2016deep}; velocities under an implicit control policy~\citep{DBLP:conf/nips/MordatchLAPT15}; or provide abstract actions~\citep{DBLP:journals/corr/HausknechtS15a}. Our learning procedures are based on prior work using actor-critic approaches with positive temporal difference updates~\citep{vanHasselt2012}.
                  
Work in biomechanics has long recognized the embodied nature of the control problem and the view that
musculotendon systems provide {\em ``preflexes''}~\citep{loeb1995control} that effectively provide a form 
intelligence by mechanics~\citep{blickhan2007intelligence}, as well as allowing for energy storage.
The control strategies for physics-based character simulations in computer animation also use all the forms of actuation
that we evaluate in this paper. Representative examples include quadratic programs that solve for joint torques~\citep{deLasa2010}, 
joint velocities for skilled bicycle stunts~\citep{Tan:2014}, muscle models 
for locomotion~\citep{Wang12optimizinglocomotion,2013-TOG-MuscleBasedBipeds},
mixed use of feed-forward torques and joint target angles~\citep{2011-TOG-quadruped}, and
joint target angles computed by learned linear (time-indexed) feedback strategies~\citep{2016-TOG-controlGraphs}.
Lastly, control methods in robotics use a mix of actuation types, including direct-drive torques (or their virtualized equivalents),
series elastic actuators, PD control, and velocity control. These methods often rely heavily on model-based solutions and thus we 
do not describe these in further detail here.

\section{Conclusions}

Our experiments suggest that action parameterizations that include basic local feedback, such as PD target angles,
MTU activations, or target velocities, can improve policy performance and learning
speed across different motions and character morphologies. Such models more accurately reflect the embodied nature of control in biomechanical
systems, and the role of mechanical components in shaping the overall dynamics of motions and their
control.  The difference between low-level and high-level action parameterizations grow with the
complexity of the characters, with high-level parameterizations scaling more gracefully to complex
characters.

Our results have only been demonstrated on planar articulated figure simulations; the extension to
3D currently remains as future work. 
Tuning actuator parameters for complex actuation models such as MTUs remains
challenging. Though our actuator optimization technique is able to improve performance as compared
to manual tuning, the resulting parameters may still not be optimal for the desired task. Therefore,
our comparisons of MTUs to other action parameterizations may not be reflective of the full
potential of MTUs with more optimal actuator parameters. Furthermore, our actuator optimization
currently tunes parameters for a specific motion, rather than a larger suite of motions, as might be expected in nature.

To better understand the effects of different action parameterizations, we believe it will be
beneficial to replicate our experiments with other reinforcement learning algorithms and motion
control tasks. As is the case with other results in this area, hyperparameter choices can have a
significant impact on performance, and therefore it is difficult to make definitive statements
with regards to the merits of the various actions spaces that we have explored. However, we believe
that the general trends we observed are likely to generalize.

Finally, it is reasonable to expect that evolutionary processes would result in the effective
co-design of actuation mechanics and control capabilities. Developing optimization and learning algorithms to allow
for this kind of co-design is a fascinating possibility for future work.


\clearpage
\bibliography{imitation}
\bibliographystyle{iclr2017_conference}
\clearpage


\section*{Supplementary Material}


\begin{algorithm}[h!]
\caption{Actor-critic Learning Using Positive Temporal Differences}
\label{alg:CACLA}
\begin{algorithmic}[1]
\STATE{$\theta \leftarrow$ random weights}
\STATE{$\phi \leftarrow$ random weights}

\item[]
\WHILE{not done}
	\FOR{$step = 1,...,m$}
		\STATE{$s \leftarrow$ start state}
		\STATE{$\lambda \leftarrow \mathrm{Ber}(\epsilon_t)$}
		\STATE{$a \leftarrow  \mu_\theta(s) + \lambda \mathcal{N}(0, \Sigma)$}
        
		\STATE Apply $a$ and simulate forward 1 step
		\STATE{$s' \leftarrow$ end state}
		\STATE{$r \leftarrow$ reward}
		\STATE{$\tau \leftarrow (s, a, r, s', \lambda)$}
	
		\STATE{store $\tau$ in replay memory}
        
        \item[]
        \IF {episode terminated}
			\STATE{Sample $s_0$ from $p_0(s)$}
            \STATE{Reinitialize state $s$ to $s_0$}
		\ENDIF
	\ENDFOR

	\item[]
	\STATE{Update critic:}
	\STATE{Sample minibatch of $n$ tuples $\{\tau_i = (s_i, a_i, r_i, \lambda_i, s_i')\}$ from replay memory}
	\FOR{each $\tau_i$}
		\STATE{$\delta_i \leftarrow r_i + \gamma V_\phi(s'_i) - V_\phi(s_i) $}
		\STATE{$\phi \leftarrow \phi + \alpha_V \frac{1}{n} \delta_i \triangledown_{\phi}V_\phi(s_i) $}
	\ENDFOR
	
	\item[]
	\STATE{Update actor:}
	\STATE{Sample minibatch of $n$ tuples $\{\tau_j = (s_j, a_j, r_j, \lambda_j, s_j')\}$ from replay memory where $\lambda_j = 1$}
	\FOR{each $\tau_j$}
		\STATE{$\delta_j \leftarrow r_j + \gamma V_\phi(s'_j) - V_\phi(s_j)$}
		\IF {$\delta_j >  0$}
        	\STATE{$\triangledown a_j \leftarrow a_j - \mu_\theta(s_j)$}
        	\STATE{$\triangledown \tilde{a}_j \leftarrow \mathrm{BoundActionGradient}					(\triangledown a_j, \mu_\theta(s_j))$}
			\STATE{$\theta \leftarrow \theta + \alpha_\pi \frac{1}{n} \triangledown_{\theta} \mu_\theta(s_j) \Sigma^{-1} \triangledown \tilde{a}_j$}
		\ENDIF
	\ENDFOR
\ENDWHILE
\end{algorithmic}
\end{algorithm}


\begin{algorithm}[h!]
\caption{Alternating Actuator Optimization}
\label{alg:mtuOpt}
\begin{algorithmic}[1]
\STATE{$\theta \leftarrow \theta_0$}
\STATE{$\psi \leftarrow \psi_0$}

\WHILE{not done}
    \STATE{$\theta \leftarrow \mathop{\mathrm{arg max}}_{\theta'} J(\pi_{\theta'}, \psi)$ with Algorithm \ref{alg:CACLA}}
    \STATE{$\psi \leftarrow \mathop{\mathrm{arg max}}_{\psi'} J(\pi_{\theta}, \psi')$ with CMA}
\ENDWHILE
\end{algorithmic}
\end{algorithm}

\clearpage


\noindent {\Large \bf MTU Actuator Optimization}

The actuator parameters $\psi$ can be interpreted as a parameterization of the dynamics of the
system $p(s' | s, a, \psi)$. The expected cumulative reward can then be re-parameterized according
to

\[ J(\pi_\theta, \psi) = \int_\mathcal{S} d_\theta(s|\psi) \int_\mathcal{A} \pi_\theta(s, a) A(s, a) da \ ds \]

where $d_\theta(s|\psi) = \int_\mathcal{S} \sum_{t=0}^T \gamma^t p_0(s_0) p(s_0 \rightarrow s | t,
\pi_\theta, \psi) ds_0$. $\theta$ and $\psi$ are then learned in tandem following Algorithm
\ref{alg:mtuOpt}. This alternating method optimizes both the control and dynamics in order to
maximize the expected value of the agent, as analogous to the role of evolution in
biomechanics. During each pass, the policy parameters $\theta$ are trained to improve the agent's
expected value for a fixed set of actuator parameters $\psi$. Next, $\psi$ is optimized using CMA to
improve performance while keeping $\theta$ fixed. The expected value of each CMA sample of $\psi$ is
estimated using the average cumulative reward over multiple rollouts.

\begin{figure}[h!]
\begin{centering}
\subfigure{   \includegraphics[width=0.32\columnwidth]{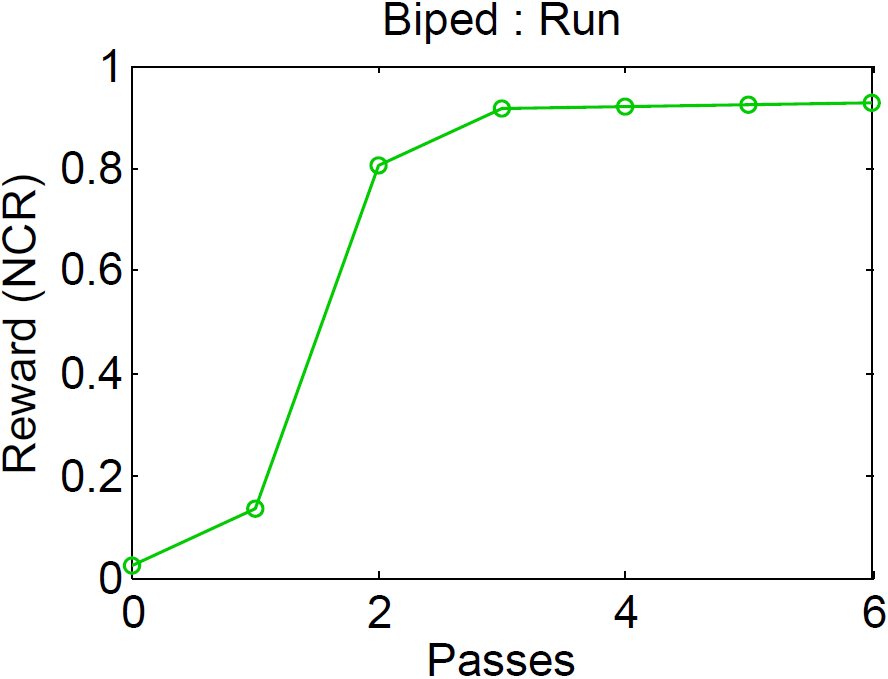}}
\subfigure{   \includegraphics[width=0.32\columnwidth]{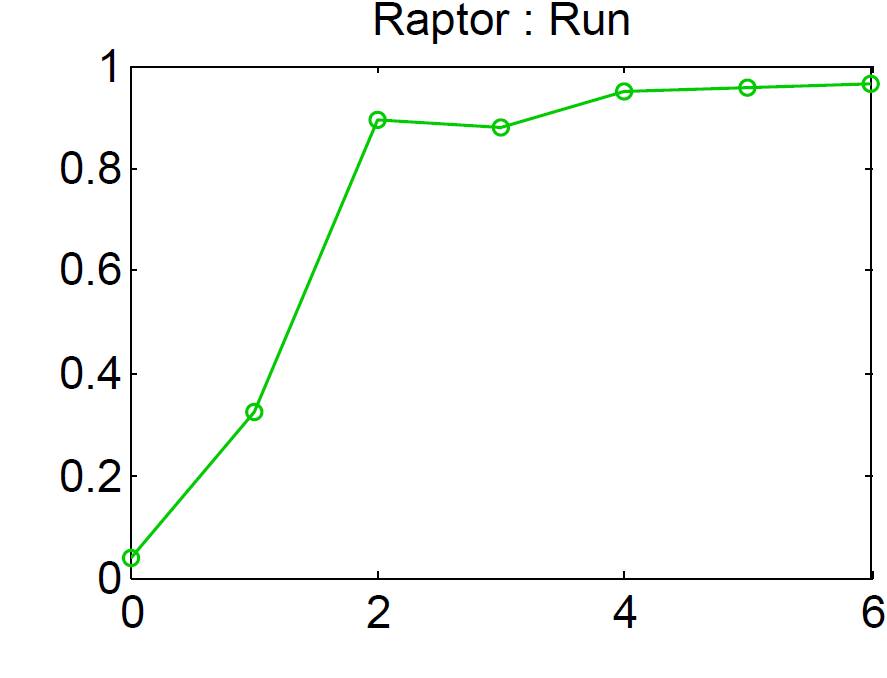}}
\subfigure{   \includegraphics[width=0.32\columnwidth]{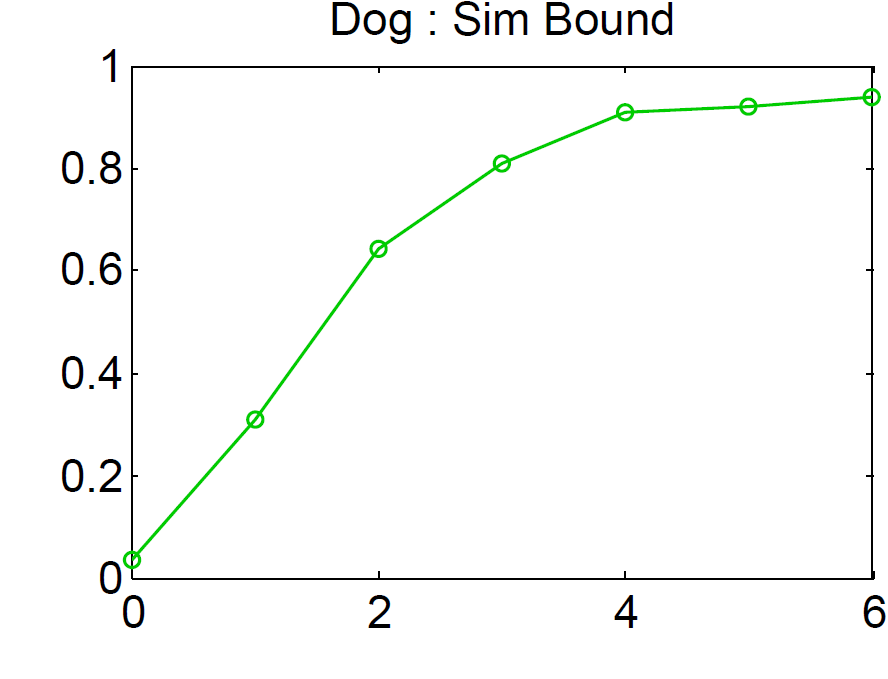}}
   \caption{Performance of intermediate MTU policies and actuator parameters per pass of actuator optimization following Algorithm \ref{alg:mtuOpt}.
     \label{fig:optMTUCurves}   }
\end{centering}
\end{figure}

\begin{figure}[h!]
\begin{centering}
\subfigure{   \includegraphics[width=0.32\columnwidth]{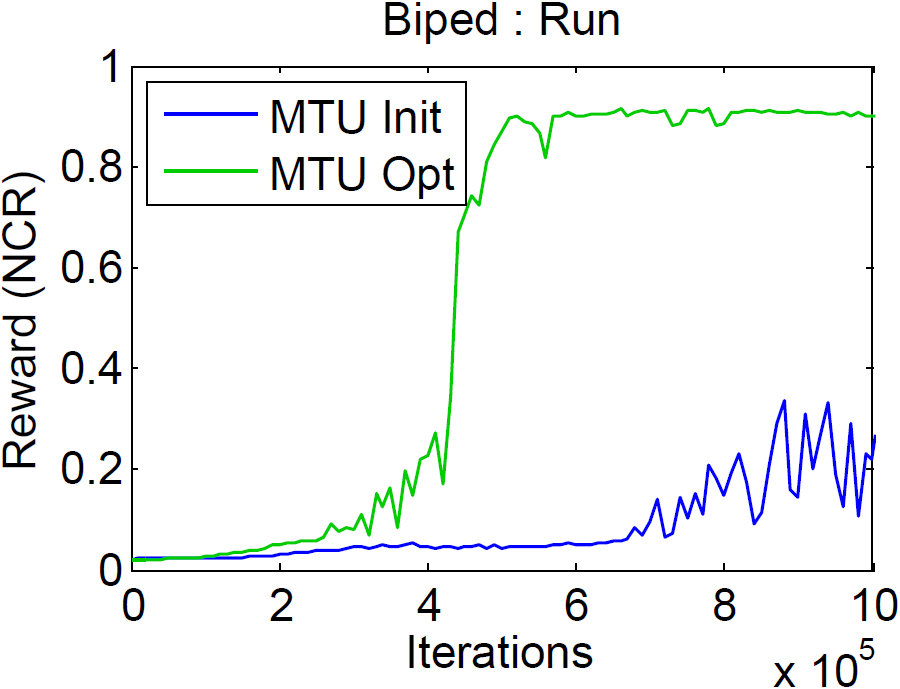}}
\subfigure{   \includegraphics[width=0.32\columnwidth]{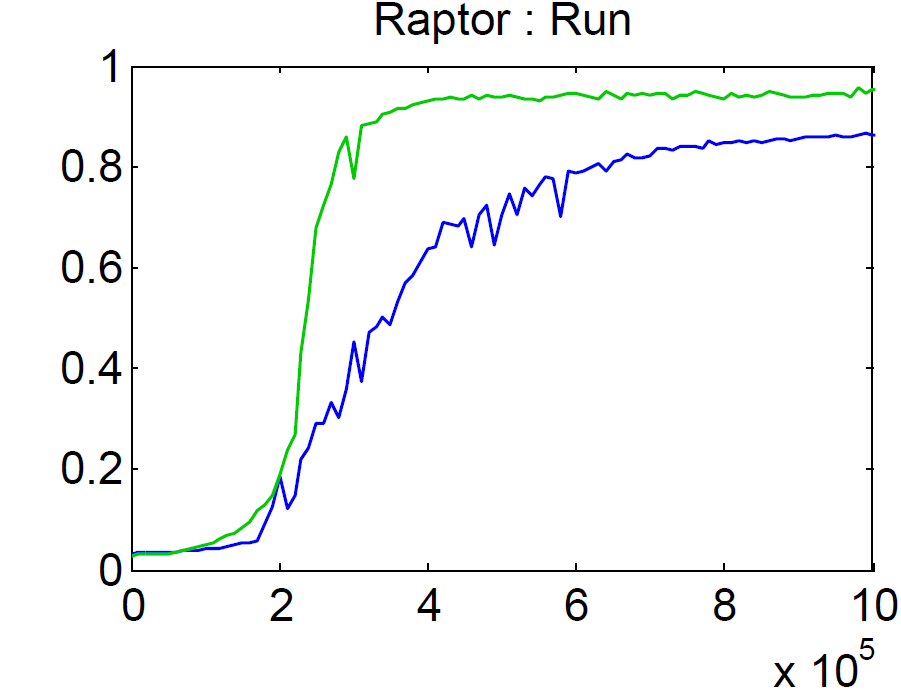}}
\subfigure{   \includegraphics[width=0.32\columnwidth]{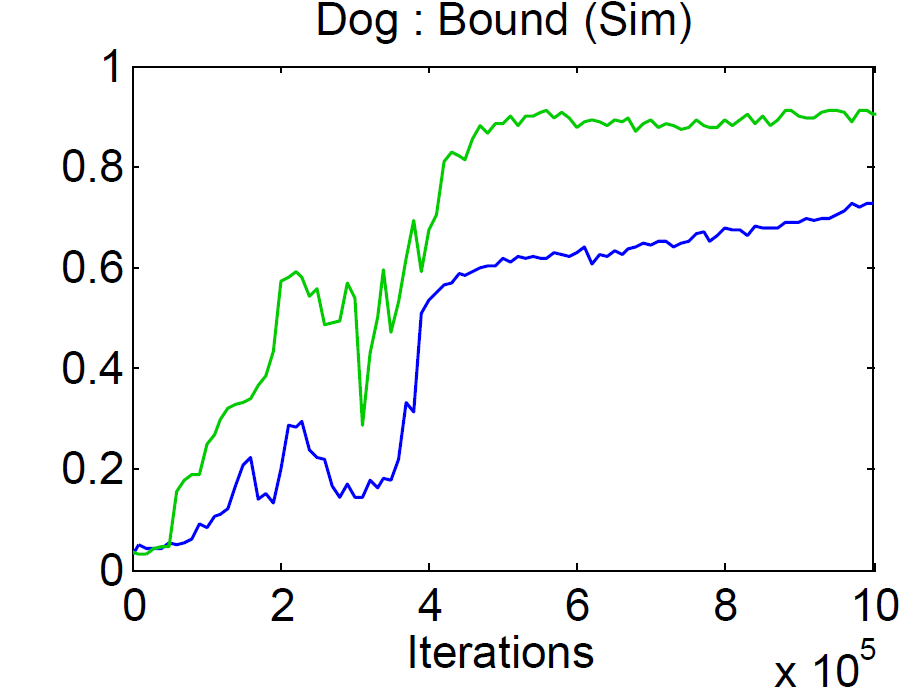}}
   \caption{Learning curves comparing initial and optimized MTU parameters.
     \label{fig:optMTUCompare}   }
\end{centering}
\end{figure}

Figure~\ref{fig:optMTUCurves} illustrates the improvement in performance during the optimization process,
as applied to motions for three different agents.
Figure~\ref{fig:optMTUCompare} compares the learning curves for the initial and final MTU parameters,
for the same three motions.


\clearpage


\noindent {\Large \bf Bounded Action Space}

Properties such as torque and neural activation limits result in bounds on the range of values that can be assumed by actions for a particular parameterization. Improper enforcement of these bounds can lead to unstable learning as the gradient information outside the bounds may not be reliable \citep{DBLP:journals/corr/HausknechtS15a}. To ensure that all actions respect their bounds, we adopt a method similar to the inverting gradients approach proposed by \citet{DBLP:journals/corr/HausknechtS15a}. Let $\triangledown a = (a - \mu(s)) A(s, a)$ be the empirical action gradient from the policy gradient estimate of a Gaussian policy. Given the lower and upper bounds $[l^i, u^i]$ of the $i$th action parameter, the bounded gradient of the $i$th action parameter $\triangledown \tilde{a}^i$ is determined according to

\[ \triangledown \tilde{a}^i = 
\begin{cases}
	l^i - \mu^i(s), 		& \text{$\mu^i(s) < l^i$ and $\triangledown a^i < 0$} \\
    u^i - \mu^i(s), 		& \text{$\mu^i(s) > u^i$ and $\triangledown a^i > 0$} \\
    \triangledown a^i,      & \text{otherwise}
\end{cases} \]

Unlike the inverting gradients approach, which scales all gradients depending on proximity to the bounds, this method preserves the empirical gradients when bounds are respected, and alters the gradients only when bounds are violated.
\\
\\

\noindent {\Large \bf Reward}

The terms of the reward function are defined as follows:
\[r_{pose} = \mathrm{exp}\left( -||q^* - q||^2_W \right), \qquad r_{vel} = \mathrm{exp}\left( -||\dot{q}^* - \dot{q}||^2_W \right)\]
\[r_{end} = \mathrm{exp} \left(-40 \sum_e ||x^*_e - x_e||^2 \right)\]
\[r_{root} = \mathrm{exp}\left( -10 (h^*_{root} - h_{root})^2 \right), \qquad r_{com} = \mathrm{exp}\left( -10 ||\dot{x}^*_{com} - \dot{x}_{com}||^2 \right) \]

$q$ and $q^*$ denotes the character pose and reference pose represented in reduced-coordinates, while $\dot{q}$ and $\dot{q}^*$ are the respective joints velocities. $W$ is a manually-specified per joint diagonal weighting matrix. $h_{root}$ is the height of the root from the ground, and $\dot{x}_{com}$ is the center of mass velocity.

\clearpage




\begin{figure}[tbh]
\begin{centering}
\subfigure{   \includegraphics[width=0.35\columnwidth]{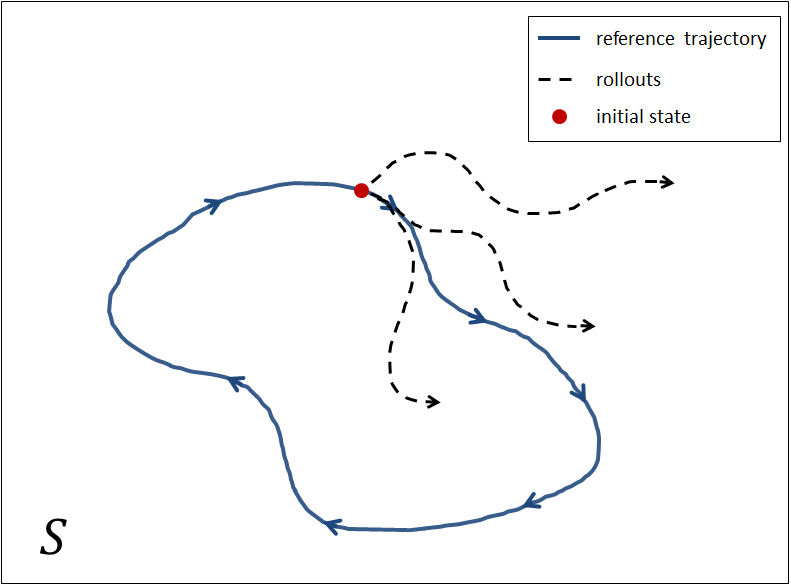}} \hspace{0.1\columnwidth}
\subfigure{   \includegraphics[width=0.35\columnwidth]{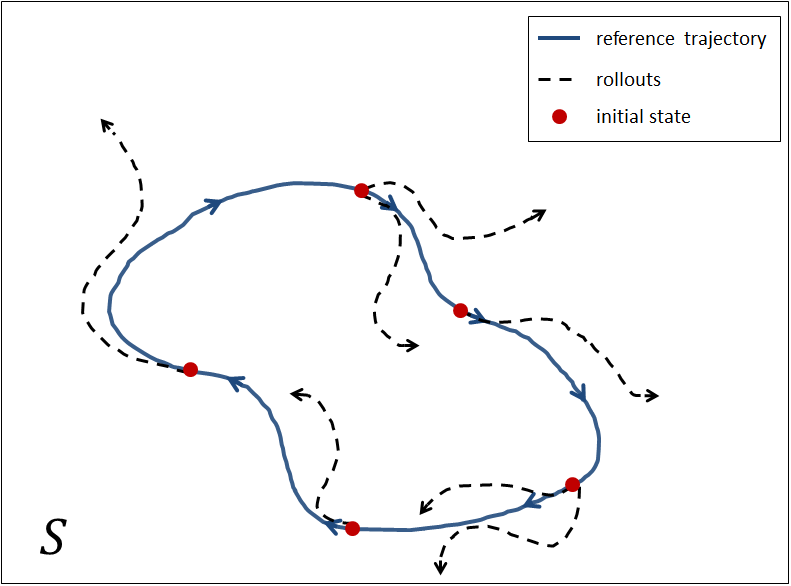}}\\
   \caption{\textbf{Left:} fixed initial state biases agent to regions of the state space near the initial state, particular during early iterations of training. \textbf{Right:} initial states sampled from reference trajectory allows agent to explore state space more uniformly around reference trajectory.
     \label{fig:initState}   }
\end{centering}
\end{figure}


\begin{table}[h!]
{ \centering  
\begin{tabular}{|c|c|c|}
\hline
Parameter & Value & Description \\ \hline
$\gamma$ &0.9   & cumulative reward discount factor\\
$\alpha_\pi$ &  0.001 & actor learning rate\\
$\alpha_V$ & 0.01   & critic learning rate\\
momentum & 0.9   & stochastic gradient descent momentum\\
$\phi$ weight decay & 0   & L2 regularizer for critic parameters\\
$\theta$ weight decay & 0.0005   & L2 regularizer for actor parameters\\
minibatch size & 32 & tuples per stochastic gradient descent step \\
replay memory size & 500000 & number of the most recent tuples stored for future updates \\
\hline
\end{tabular} \\
}
\caption{Training hyperparameters.}
\label{tab:hyperParams}
\end{table}

\clearpage


\begin{table}[h]
{ \centering  
\begin{tabular}{|c|c|c|c|}
\hline
Character + Actuation Model & State Parameters & Action Parameters & Actuator Parameters \\ \hline
Biped + Tor & 58 & 6 & 0 \\
Biped + Vel & 58 & 6 & 6 \\
Biped + PD & 58 & 6 & 12 \\
Biped + MTU & 74 & 16 & 114 \\
\hline
Raptor + Tor & 154 & 18 & 0 \\
Raptor + Vel & 154 & 18 & 18 \\
Raptor + PD & 154 & 18 & 36 \\
Raptor + MTU & 194 & 40 & 258 \\
\hline
Dog + Tor & 170 & 20 & 0 \\
Dog + Vel & 170 & 20 & 20 \\
Dog + PD & 170 & 20 & 40 \\
Dog + MTU & 214 & 44 & 282 \\
\hline
\end{tabular} \\
}
\caption{The number of state, action, and actuation model parameters for different characters and actuation models.}
\label{tab:spaceDims}
\end{table}


\begin{table}[h!!]
{ \centering  
\begin{tabular}{|c|c|c|c|}
\hline
Character + Actuation & Motion & Performance (NCR) & Learning Speed (AUC) \\ \hline
Biped + Tor & Walk & 0.7662 $\pm$ 0.3117 & 0.4788 \\
Biped + Vel & Walk & 0.9520 $\pm$ 0.0034 & 0.6308 \\
Biped + PD & Walk & 0.9524 $\pm$ 0.0034 & 0.6997 \\
Biped + MTU & Walk & \textbf{0.9584 $\pm$ 0.0065} & \textbf{0.7165} \\
\hline
Biped + Tor & March & 0.9353 $\pm$ 0.0072 & 0.7478 \\
Biped + Vel & March & \textbf{0.9784 $\pm$ 0.0018} & 0.9035 \\
Biped + PD & March & 0.9767 $\pm$ 0.0068 & \textbf{0.9136} \\
Biped + MTU & March & 0.9484 $\pm$ 0.0021 & 0.5587 \\
\hline
Biped + Tor & Run & 0.9032 $\pm$ 0.0102 & 0.6938 \\
Biped + Vel & Run & \textbf{0.9070 $\pm$ 0.0106} & 0.7301 \\
Biped + PD & Run & 0.9057 $\pm$ 0.0056 & \textbf{0.7880} \\
Biped + MTU & Run & 0.8988 $\pm$ 0.0094 & 0.5360 \\
\hline
Raptor + Tor & Run (Sim) & 0.7265 $\pm$ 0.0037 & 0.5061 \\
Raptor + Vel & Run (Sim) & 0.9612 $\pm$ 0.0055 & 0.8118 \\
Raptor + PD & Run (Sim) & \textbf{0.9863 $\pm$ 0.0017} & \textbf{0.9282} \\
Raptor + MTU & Run (Sim) & 0.9708 $\pm$ 0.0023 & 0.6330 \\
\hline
Raptor + Tor & Run & 0.6141 $\pm$ 0.0091 & 0.3814 \\
Raptor + Vel & Run & 0.8732 $\pm$ 0.0037 & 0.7008 \\
Raptor + PD & Run & \textbf{0.9548 $\pm$ 0.0010} & \textbf{0.8372} \\
Raptor + MTU & Run & 0.9533 $\pm$ 0.0015 & 0.7258 \\
\hline
Dog + Tor & Bound (Sim) & 0.8016 $\pm$ 0.0034 & 0.5472 \\
Dog + Vel & Bound (Sim) & 0.9788 $\pm$ 0.0044 & 0.7862 \\
Dog + PD & Bound (Sim) & \textbf{0.9797 $\pm$  0.0012} & \textbf{0.9280} \\
Dog + MTU & Bound (Sim) & 0.9033 $\pm$ 0.0029 & 0.6825 \\
\hline
Dog + Tor & Rear-Up & 0.8151 $\pm$ 0.0113 & 0.5550 \\
Dog + Vel & Rear-Up & 0.7364 $\pm$ 0.2707 & 0.7454 \\
Dog + PD & Rear-Up & \textbf{0.9565 $\pm$ 0.0058} & \textbf{0.8701} \\
Dog + MTU & Rear-Up & 0.8744 $\pm$ 0.2566 & 0.7932 \\
\hline
\end{tabular} \\
}
\caption{Performance of policies trained for the various characters and actuation models. Performance is measured using the normalized cumulative reward (NCR) and learning speed is represented by the normalized area under each learning curve (AUC). The best performing parameterizations for each character and motion are in bold.}
\label{tab:perf}
\end{table}

\clearpage


\begin{figure}[tbh]
\begin{centering}
\subfigure{   \includegraphics[width=0.7\columnwidth]{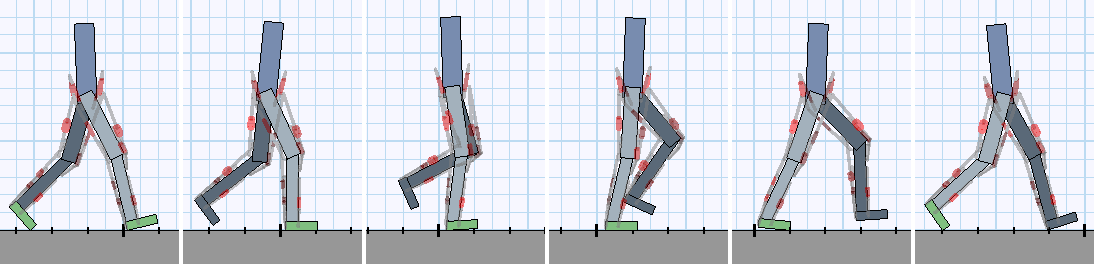}}
\subfigure{   \includegraphics[width=0.7\columnwidth]{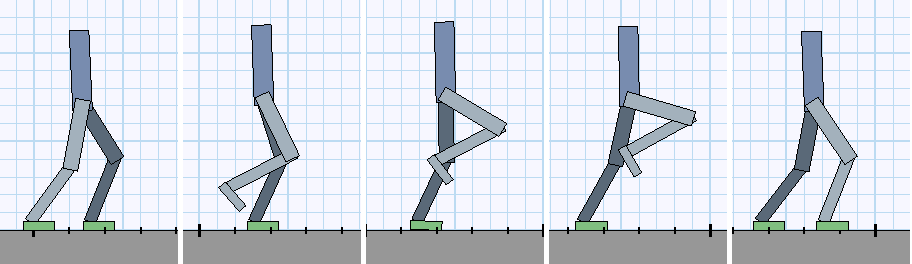}}
\subfigure{   \includegraphics[width=0.7\columnwidth]{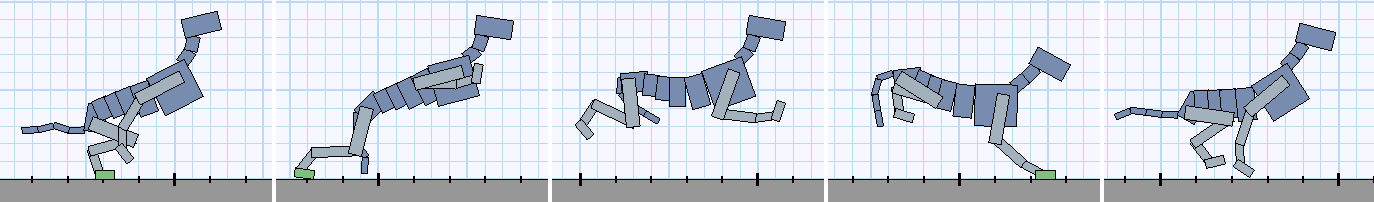}}
\subfigure{   \includegraphics[width=0.7\columnwidth]{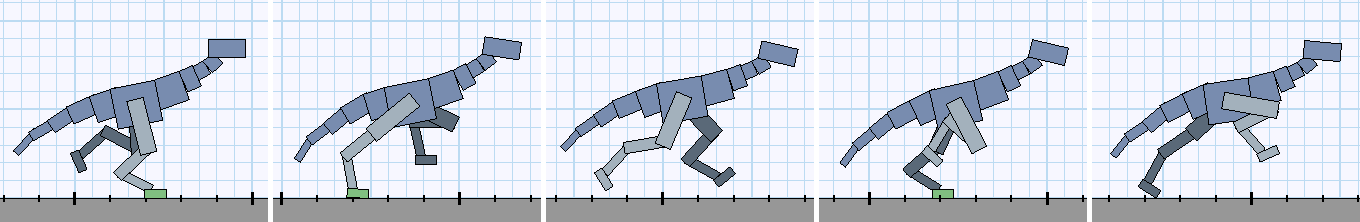}}
   \caption{Simulated Motions Using the PD Action Representation. The top row uses an MTU action space
    while the remainder are driven by a PD action space.
          \label{fig:motions}   }
\end{centering}
\end{figure}


\clearpage


\end{document}